%% file: iclr2027_conference.tex
\documentclass{article}

\usepackage{iclr2027_conference,times}
\input{math_commands.tex}

\usepackage[table]{xcolor}
\usepackage{hyperref}
\usepackage{url}
\usepackage{xspace}
\usepackage{booktabs}
\usepackage{array}
\usepackage{longtable}
\usepackage{graphicx}
\usepackage{amssymb}

\newcommand{\algname}{\textsc{SHarP}\xspace}

\title{\algname: Saliency-based Pruning of Agent Harnesses}

\author{Xinyi Gao$^1$,
  Qiucheng Wu$^1$,
  Kaizhi Qian$^2$,
  Handong Zhao$^3$,
  Shiyu Chang$^1$,
  Yang Zhang$^2$ \\[0.5em]
  $^1$University of California, Santa Barbara \\
  $^2$MIT-IBM Watson AI Lab \\ 
  $^3$Adobe Research \\[0.5em]
  \texttt{xinyigao@ucsb.edu}
}

\iclrfinalcopy 
\begin{document}
\maketitle
\lhead{Preprint.}

\input{sections/abstract}

\renewcommand{\thefootnote}{}
\footnotetext{Code available at \url{https://github.com/UCSB-NLP-Chang/SHarP}}
\renewcommand{\thefootnote}{\arabic{footnote}}

\input{sections/introduction}

\input{sections/related_work}

\input{sections/methods}

\input{sections/experiments}

\input{sections/conclusion}

\input{sections/acknowledgments}

\input{sections/reproducibility}

\input{sections/ai_use}

\bibliography{iclr2027_conference}
\bibliographystyle{iclr2027_conference}

\appendix
\input{sections/appendix}

\end{document}

%% file: math_commands.tex
\usepackage{amsmath,amsfonts,bm}

\def\eqref#1{equation~\ref{#1}}

\def\1{\bm{1}}

\DeclareMathAlphabet{\mathsfit}{\encodingdefault}{\sfdefault}{m}{sl}
\SetMathAlphabet{\mathsfit}{bold}{\encodingdefault}{\sfdefault}{bx}{n}



%% file: sections/abstract.tex
\begin{abstract}
Agent harnesses are systems that coordinate model calls, tool use, and task execution to help large language models complete complex tasks. To meet task requirements and address failures, these systems are often iteratively refined by amending and patching their instructions, tools, and workflows, continuously increasing harness complexity. It is therefore unclear whether some resulting harness modules are redundant, introducing substantial token overhead with little, if any, performance gain. Inspired by neural network pruning, in this paper, we study \textbf{harness pruning} as a means of striking a better balance between task performance and token cost. We propose \algname (\textbf{S}aliency-based \textbf{Har}ness \textbf{P}runing), a simple yet effective pruning strategy based on the saliency of each harness module with respect to performance and efficiency. Specifically, we first identify tools, instructions, and supporting mechanisms as components that can be individually disabled. We then estimate the saliency of each module by ablating it and assessing its task performance and token cost relative to the full set of single-module ablations. Modules with the smallest contribution to performance or largest computational overhead are subsequently pruned. Our evaluation across various harnesses on held-out validation sets reveals a surprising finding: most harnesses that we studied are highly redundant and can maintain comparable performance and efficiency even after a substantial portion of their modules are pruned. Our pruning approach and empirical findings provide new perspectives on agent harness design and optimization.
\end{abstract}

%% file: sections/introduction.tex
\section{Introduction}
\label{sec:introduction}

Large language model agents rely on more than the model itself to complete complex tasks. In practice, an agent harness provides the tools, instructions, and execution logic that allow the model to interact with its environment and work through a task \citep{grace2026demystifying,trivedy2026anatomy}. The way these elements are organized, including which tools are available and how context and feedback are managed, can influence how effectively the agent completes its tasks. Harness design therefore offers a way to improve agent performance without changing the underlying model. With this goal, research and engineering efforts have explored how to adapt harnesses to the needs of particular tasks and applications \citep{lee2026meta,lin2026agentic,zhang2026self}.

Harnesses often undergo iterative adaptation and refinement to optimize performance under evolving task requirements. Such adaptation typically involves amending and patching existing harnesses, often by adding new instructions, tools, in-context examples, and processing steps. As a result, harnesses tend to grow monotonically in size and complexity, leading to increased computational and token costs. Some recent work has begun to recognize the problem of harness bloat. \citet{sengupta2026harbor} find that adding mechanisms to an agent harness does not necessarily improve task performance, highlighting the need to consider their benefits and costs together. \citet{agostino2026herding} explore limiting agents' access to context and tools as a means of reducing harness complexity. Yet, there remains no systematic way to quantify how much redundancy exists in an agent harness or to effectively remove redundant components.

Neural network pruning offers a natural source of inspiration. In a common pruning paradigm, neural network weights are ranked by a saliency score, and those with the lowest saliency are removed (set to zero). Studies have found that even simple heuristics for estimating saliency, such as weight magnitude \citep{han2015learning} or gradient magnitude \citep{NEURIPS2019_2c601ad9}, can be highly effective: neural networks can retain comparable performance, or even achieve slight performance gains, after a substantial portion of their weights are pruned \citep{frankle2018lottery}. This raises two natural questions: \textbf{Can agent harnesses likewise be substantially pruned with little or no loss in performance? Can similarly simple heuristics be used to effectively prune agent harnesses?}

However, pruning agent harnesses introduces two key challenges. \textit{First}, unlike neural network pruning, where removing a weight generally improves computational efficiency, pruning a harness component does not necessarily reduce computational or token costs. For example, removing a module that helps the agent recognize task completion and terminate execution may cause it to continue making model calls after the task is already complete, increasing token usage. Therefore, deciding which components to prune requires jointly considering their impact on both task performance and efficiency. \textit{Second}, evaluating an agent harness is itself costly in both time and inference resources, making it impractical to exhaustively evaluate different combinations of pruned components. A practical harness pruning method must therefore be highly evaluation-efficient.

\begin{figure}[t]
\centering
\includegraphics[width=\linewidth]{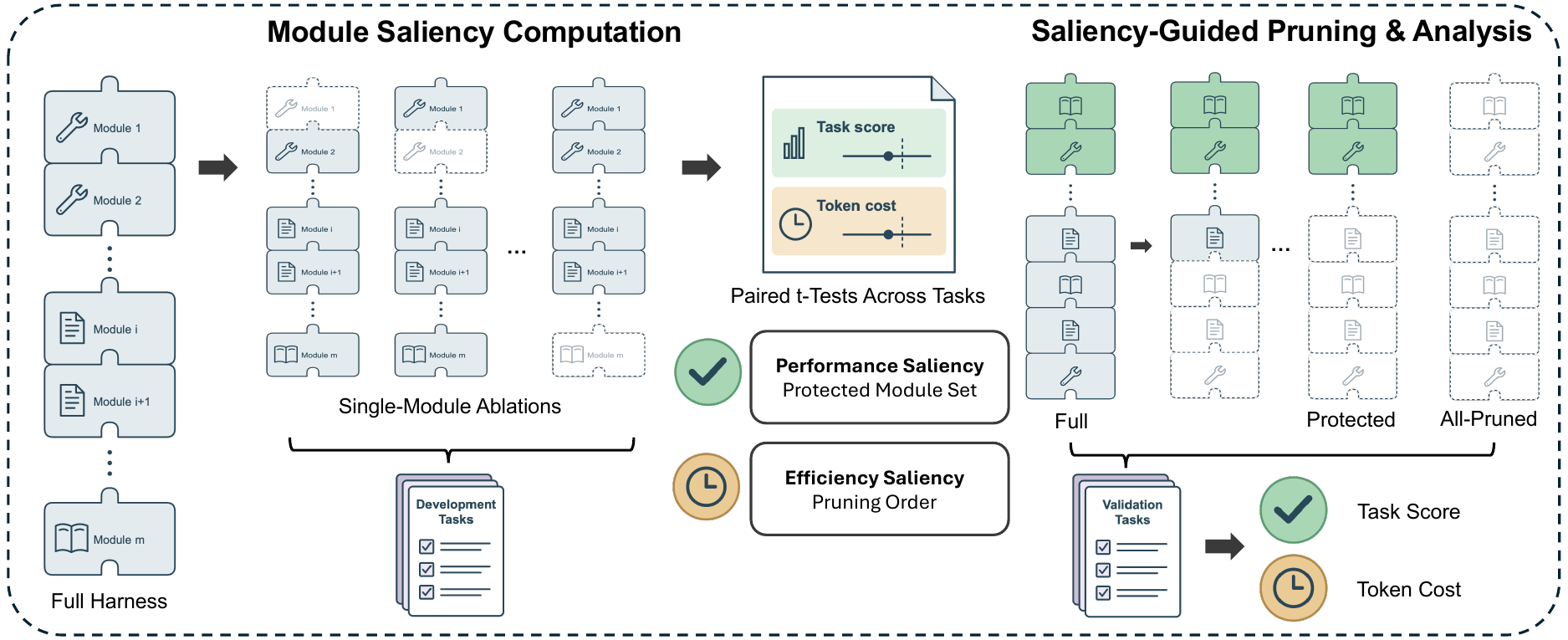}
\vspace{-5mm}
\caption{Overview of \algname's efficiency-oriented pruning. Single-module ablations estimate performance and efficiency saliency, which determine the protected module set (green) and pruning order, respectively. Pruned configurations and an all-pruned baseline are evaluated for task performance and token cost.}
\label{fig:pipeline}
\vspace{-5mm}
\end{figure}

To address these challenges, we propose \algname (\textbf{S}aliency-based \textbf{Har}ness \textbf{P}runing), a simple pruning strategy for agent harnesses based on two dimensions of saliency: performance saliency and efficiency saliency. Figure~\ref{fig:pipeline} summarizes the pipeline. We first identify tools, instructions, and supporting mechanisms as modules that can be individually disabled. For each prunable module, we estimate its performance and efficiency saliency through an evaluation-efficient single-module ablation: we disable the module and measure the resulting changes in task performance and token cost, requiring only one additional evaluation per module. Based on the resulting saliency scores, we propose two pruning strategies: efficiency-oriented pruning, which prunes modules with the least favorable efficiency saliency while protecting those with a high impact on performance, and performance-oriented pruning, which prunes modules with the least favorable performance saliency while protecting those with a high impact on efficiency.

Our evaluation across four agent harnesses on their respective validation sets reveals surprisingly favorable performance--efficiency tradeoffs: all four harnesses maintain comparable performance while achieving at least a 10\% reduction in token cost. Two of the harnesses even improve task performance while reducing token cost through pruning. These results reveal substantial redundancy in existing agent harnesses and demonstrate that simple saliency-based pruning can effectively exploit this redundancy, providing new perspectives on agent harness design and optimization.

%% file: sections/related_work.tex
\section{Related Work}
\paragraph{Harness in agentic workflow.} Harnesses provide the tools, instructions, and execution mechanisms that enable language models to perform long-horizon, multi-step tasks~\citep{zhou2026externalization,pan2026natural,ning2026code}.
Work on harness construction searches over executable programs to find effective combinations of reasoning and tool-use operations~\citep{zhang2025aflow,hu2025automated}. 
To refine an existing harness, other methods use execution traces and task feedback to revise its code and components~\citep{lee2026meta,lin2026agentic,zhang2026self}. 
For particular tasks and environments, recent approaches adapt the runtime interface based on recurring failures~\citep{xu2026adapting} or generate a specific harness for each task~\citep{zhang2026jitagentscalingharnessintelligence}.

\paragraph{Harness efficiency.} Executing multi-step workflows can require repeated model calls and increasingly long contexts, making efficiency an important consideration in harness development.
Several studies address this overhead by restricting context and tool exposure according to task roles~\citep{agostino2026herding,liu2025rcr} or by compressing the interaction history presented to the model~\citep{wu2025resum,ye2025agentfold}.
However, compressing the interaction history can introduce overhead.
Recent work compares model-generated summaries with simply omitting older tool observations, finding comparable task performance while reducing execution cost~\citep{lindenbauer2025complexity}.
Further, recent work searches over harness configurations while accounting for task performance and cost.
\citet{sengupta2026harbor} formulate this search as constrained optimization under noisy task evaluations, accounting for both the expense of configuration search and the cost of deploying the resulting harness.
Other work searches over prompts and tools, reporting held-out performance alongside cost and reliability measures~\citep{zhao2026beyond}.

\paragraph{Pruning.} Pruning reduces system complexity by identifying components that can be removed while preserving performance. 
In neural networks, existing methods have used weight magnitude and task-specific connection sensitivity to select individual parameters for removal~\citep{han2015learning,lee2018snip}. 
Since removing individual weights does not necessarily reduce execution cost, structured pruning instead targets larger components in models such as filters, channels, and layers~\citep{wen2016learning,li2016pruning,liu2017learning}. 
More recently, studies have examined whether removing redundant components can improve the efficiency of agentic workflows.
Work on multi-agent communication removes redundant messages or communication links to reduce token use~\citep{wang2025agentdropout,zhang2025cut}. 
Other approaches also adjust the number of agents, either by jointly pruning agents and communication links or by removing and replacing costly agents~\citep{li2025adaptive,chen2026agentslimming}. 
Closer to harnesses, controlled module ablations show that additional execution structure does not consistently improve task outcomes~\citep{pan2026natural,lin2026agentic}. 
However, these studies leave open how to prune heterogeneous components of an existing harness, including tools, instructions, and supporting mechanisms, based on their contributions to both task performance and token cost. 
In this work, we use single-module ablations to construct a sequence of pruned harness configurations and evaluate that sequence on held-out tasks.

%% file: sections/methods.tex
\section{Methods}
\label{sec:methods}

Given an agent harness, \algname aims to assess and reduce its redundancy by pruning its modules. This process involves four steps:

\begin{list}{\textbullet}{
  \setlength{\leftmargin}{1.2em}
  \setlength{\labelwidth}{0.5em}
  \setlength{\labelsep}{0.7em}
  \setlength{\itemsep}{3pt}
  \setlength{\parsep}{0pt}
  \setlength{\topsep}{4pt}
  \setlength{\partopsep}{0pt}
}
\item \textbf{Module identification:} Identify modules that can be individually removed;
\item \textbf{Saliency computation:} For each prunable module, we compute its performance and efficiency saliency on a development set;
\item \textbf{Saliency-guided pruning:} We remove a subset of modules based on the efficiency-oriented or the performance-oriented rule;
\item \textbf{Pruning evaluation:} We evaluate the impact of pruning on a validation set.
\end{list}

The following subsections describe these four steps in detail, respectively.

\subsection{Module identification and harness configurations}
\label{sec:method-notation}

Unlike neural network pruning, agent harnesses do not have predefined pruning units. In this step, we manually partition the harnesses into prunable modules. The general partition rule is to follow existing implementation boundaries, such as a registered tool, an instruction block identified by its heading or tags, a retrieved skill, or a distinct processing mechanism. In addition, highly coupled elements in a harness, such as a tool and its accompanying usage instructions, should be partitioned into a single module.

For each module, we also need to define what it means to prune it, and what the resulting harness would look like. For neural networks, pruning a weight means setting the weight to zero. For agent harnesses, pruning a module may involve different forms. For example, pruning a search tool involves removing the tool and its instructions from the prompt, and disabling the agents' access to the tool. On the other hand, pruning a history summarization mechanism involves always keeping the full history without summarizing it. 

For ease of subsequent discussion, we introduce some notation. Let $\mathcal{C}=\{1,\ldots,m\}$ index the candidate modules. For a retained set $\mathcal{S}\subseteq\mathcal{C}$, we define $\mathcal{A}(\mathcal{S})$ as the harness that retains the modules in $\mathcal{S}$ and disables the remaining candidates. Thus, $\mathcal{A}(\mathcal{C})$ is the full harness and $\mathcal{A}(\varnothing)$ is the all-pruned configuration, which retains the task definition and the basic code needed to run the agent and interact with the task environment.

Appendix~\ref{sec:appendix-harness-modules} lists the candidate modules and their functions for each harness, explains special cases where disabling a module preserves part of its functionality or introduces alternative behavior, and describes the capabilities retained in the all-pruned configurations.

\subsection{Module saliency computation}
\label{sec:method-screening}

The saliency scores of each module are computed on a development set consisting of $n$ agentic tasks. For each module $i$, we disable this module and run the resulting harness, denoted by $\mathcal{A}_{-i}=\mathcal{A}(\mathcal{C}\setminus\{i\})$, on the $n$ agentic tasks. For each task $t$, we record two values:

$\bullet$ \textbf{Task score}, denoted as $y_{ti}\in[0,1]$, which is the task-dependent success metric with higher values indicating better success, such as a binary indicator of task success (see Section~\ref{sec:experimental-setup} for the metrics used for each benchmark);

$\bullet$ \textbf{Token cost}, denoted as $c_{ti}\geq 0$, which counts input and output tokens across the model calls accounted for by the harness.

Model configuration, resource budgets, and evaluation conditions are fixed across all single-module ablations within each setting.

We then compute two difference terms,
\begin{equation}
d^{\mathrm{perf}}_{ti}=y_{ti}-\widetilde y_t,
\qquad
d^{\mathrm{tok}}_{ti}=c_{ti}-\widetilde c_t,
\label{eq:paired-differences}
\end{equation}
where $\widetilde y_t$ and $\widetilde c_t$ are the medians on the same task over all the single-module-ablated harnesses, \emph{i.e.},
\begin{equation}
\widetilde y_t=\operatorname{median}\!\left(y_{t1},\ldots,y_{tm}\right),
\qquad
\widetilde c_t=\operatorname{median}\!\left(c_{t1},\ldots,c_{tm}\right).
\label{eq:field-reference}
\end{equation}
The two difference terms, $d^{\mathrm{perf}}_{ti}$ and $d^{\mathrm{tok}}_{ti}$, measure the impact of removing module $i$ on task $t$. If $d^{\mathrm{perf}}_{ti} < 0$, or $d^{\mathrm{tok}}_{ti} > 0$, it means that removing module $i$ would harm the task performance or increase token cost, respectively, beyond the median level, and thus module $i$ should be considered more important to keep.

Based on this intuition, we further define the difference averaged across all the tasks, \emph{i.e.},
\begin{equation}
    \Delta_i^{\mathrm{perf}} = \mathbb{E}_t[d^{\mathrm{perf}}_{ti}], \qquad
    \Delta_i^{\mathrm{tok}} = \mathbb{E}_t[d^{\mathrm{tok}}_{ti}],
\end{equation}
and test the one-sided statistical significance of $\Delta_i^{\mathrm{perf}}$ and $\Delta_i^{\mathrm{tok}}$ under the null hypotheses
\begin{equation}
H_{0,i}^{\mathrm{perf}}:\Delta_i^{\mathrm{perf}}\geq 0,
\qquad
H_{0,i}^{\mathrm{eff}}:\Delta_i^{\mathrm{tok}}\leq 0,
\label{eq:eff-hypotheses}
\end{equation}
We use paired t-tests and compute the $p$-values, denoted as $p_i^{\mathrm{perf}}$ and $p_i^{\mathrm{eff}}$, respectively (see Appendix~\ref{sec:appendix-saliency-tests} for test details). The smaller these $p$-values are, the more likely that the null hypotheses are rejected, and the more important module $i$ is in terms of achieving high performance/high efficiency. We use $p_i^{\mathrm{perf}}$ and $p_i^{\mathrm{eff}}$ to quantify \textbf{performance saliency} and \textbf{efficiency saliency}, respectively.

\subsection{Saliency-guided pruning}
\label{sec:method-reconstruction}

We consider two pruning strategies, \textbf{efficiency-oriented pruning} and \textbf{performance-oriented pruning}, which prioritize improving efficiency and task performance, respectively. For conciseness, we only describe efficiency-oriented pruning in this section. Performance-oriented pruning can be straightforwardly derived via symmetry.

Our efficiency-oriented pruning strategy prioritizes pruning away modules that are harmful to efficiency, while excluding modules that are critical to performance from the pruning list. Specifically, we define a protected module set, consisting of modules that are crucial for performance:
\begin{equation}
\mathcal{G}=\{i\in\mathcal{C}:p_i^{\mathrm{perf}}\leq\alpha\},
\qquad \alpha=0.05.
\label{eq:performance-gate}
\end{equation}

For the remaining unprotected modules, $\mathcal{C}\setminus \mathcal{G}$, we rank them by their efficiency saliency, \emph{i.e.}, ranking $p_i^{\mathrm{eff}}$ from high to low. Define $\mathcal{A}(\mathcal{S}_{m-k})$ as the harness with $k$ modules pruned. We obtain $\mathcal{A}(\mathcal{S}_{m-k})$ simply by removing the $k$ modules with the highest $p_i^{\mathrm{eff}}$. In this way, we can obtain pruned harnesses at different pruning budgets: $\{\mathcal{A}(\mathcal{S}_{l})\}, l\in\{|\mathcal{G}|, \cdots, m\}$. The smallest harness has the size of $|\mathcal{G}|$ because modules in the protected set cannot be pruned.

\subsection{Pruning analysis}
\label{sec:method-analysis}

We evaluate each configuration $\mathcal{A}(\mathcal{S}_k)$ described above on a held-out validation set that is disjoint from the development set. All configurations are evaluated on the same tasks under matched conditions. By default, we evaluate each configuration with $R=3$ trials per task. For $N$ validation tasks, let $y^{\mathrm{val}}_{trk}$ and $c^{\mathrm{val}}_{trk}$ denote task score and total token cost for trial $r$ of task $t$ under $\mathcal{A}(\mathcal{S}_k)$. We summarize each configuration by
\begin{equation}
\widehat P_k=\frac{1}{NR}\sum_{t=1}^{N}\sum_{r=1}^{R}y^{\mathrm{val}}_{trk},
\qquad
\widehat C_k=\frac{1}{NR}\sum_{t=1}^{N}\sum_{r=1}^{R}c^{\mathrm{val}}_{trk}.
\label{eq:evaluation-metrics}
\end{equation}
For each configuration, we report task performance and token cost, where $k$ denotes the number of retained modules. Comparing these measurements with the full harness and the all-pruned configuration shows how task performance and token cost vary with the number of pruned modules.

%% file: sections/experiments.tex
\section{Experiments}
\label{sec:experiments}

\begin{figure}[p]
\centering
\begin{minipage}[c]{0.12\linewidth}
\scriptsize
\raggedright
(a)\\OpenHands\\GAIA
\end{minipage}\hfill
\begin{minipage}[c]{0.435\linewidth}
\includegraphics[width=\linewidth]{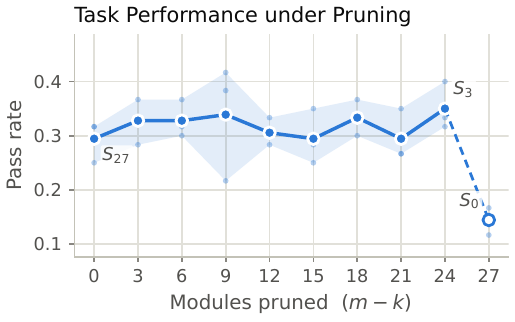}
\end{minipage}\hfill
\begin{minipage}[c]{0.435\linewidth}
\includegraphics[width=\linewidth]{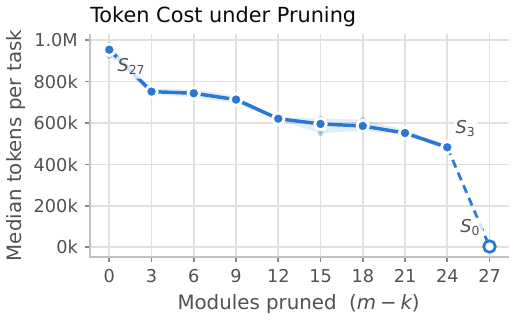}
\end{minipage}
\par\vspace{4pt}
\begin{minipage}[c]{0.12\linewidth}
\scriptsize
\raggedright
(b)\\PaperQA2\\LitQA2
\end{minipage}\hfill
\begin{minipage}[c]{0.435\linewidth}
\includegraphics[width=\linewidth]{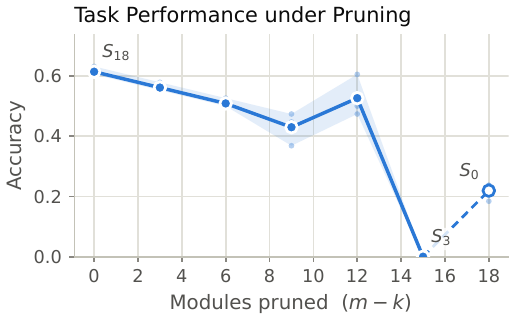}
\end{minipage}\hfill
\begin{minipage}[c]{0.435\linewidth}
\includegraphics[width=\linewidth]{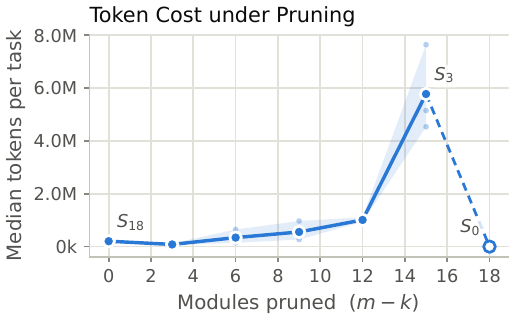}
\end{minipage}
\par\vspace{4pt}
\begin{minipage}[c]{0.12\linewidth}
\scriptsize
\raggedright
(c)\\LIFE-harness\\$\tau^2$-bench\\airline
\end{minipage}\hfill
\begin{minipage}[c]{0.435\linewidth}
\includegraphics[width=\linewidth]{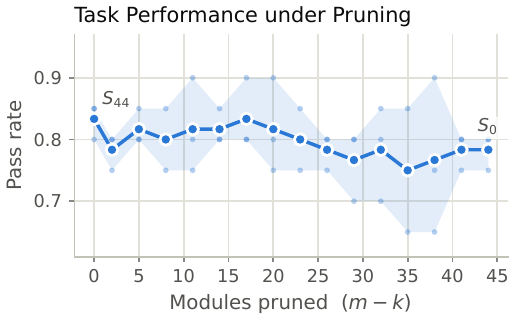}
\end{minipage}\hfill
\begin{minipage}[c]{0.435\linewidth}
\includegraphics[width=\linewidth]{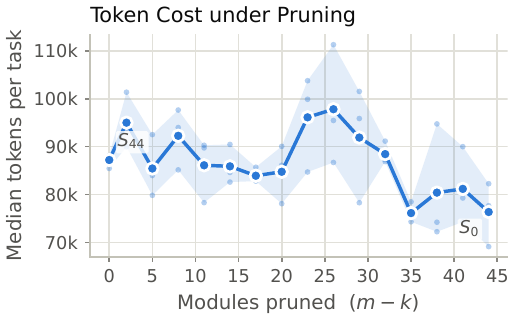}
\end{minipage}
\par\vspace{4pt}
\begin{minipage}[c]{0.12\linewidth}
\scriptsize
\raggedright
(d)\\LIFE-harness\\$\tau^2$-bench\\retail
\end{minipage}\hfill
\begin{minipage}[c]{0.435\linewidth}
\includegraphics[width=\linewidth]{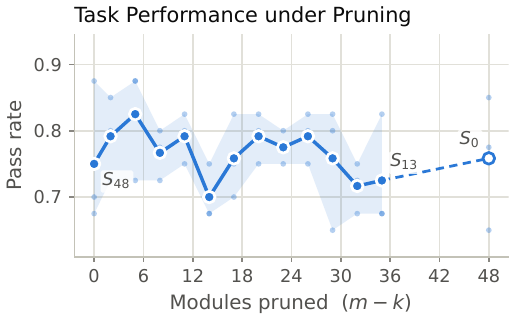}
\end{minipage}\hfill
\begin{minipage}[c]{0.435\linewidth}
\includegraphics[width=\linewidth]{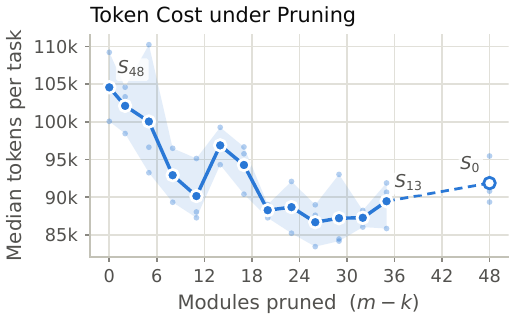}
\end{minipage}
\par\vspace{4pt}
\begin{minipage}[c]{0.12\linewidth}
\scriptsize
\raggedright
(e)\\JIT agentfold\\DeepPlanning-\\Shopping
\end{minipage}\hfill
\begin{minipage}[c]{0.435\linewidth}
\includegraphics[width=\linewidth]{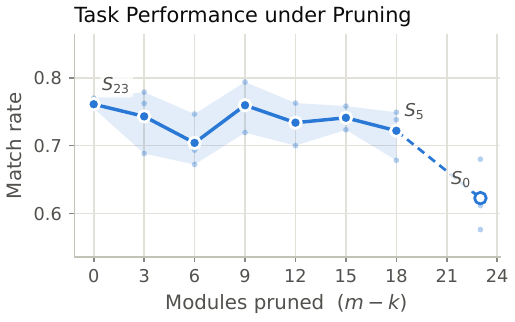}
\end{minipage}\hfill
\begin{minipage}[c]{0.435\linewidth}
\includegraphics[width=\linewidth]{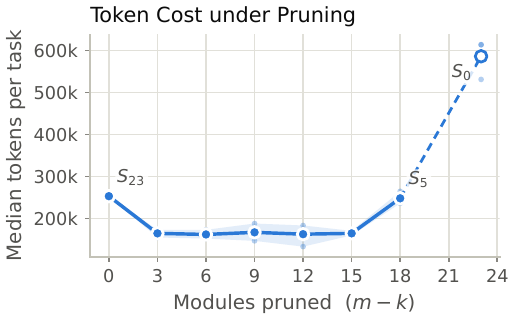}
\end{minipage}
\par\smallskip
\includegraphics[width=0.751\linewidth]{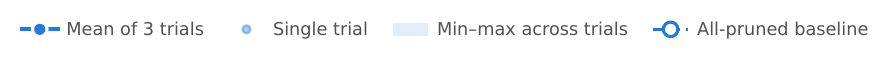}
\vspace{-3mm}
\caption{Task performance (left) and token cost (right) as more modules are pruned under the efficiency-oriented strategy. Lines show means over three trials; faint points show individual trials, and shading spans their minimum and maximum. Token cost uses the per-trial median across tasks. When the protected set is nonempty, hollow circles mark the additional all-pruned baseline.}
\label{fig:appendix-pruning-curves}
\end{figure}

\subsection{Experimental setup}
\label{sec:experimental-setup}

\paragraph{Harnesses and benchmarks.} We study five settings spanning four harnesses: OpenHands~\citep{wang2026openhandssoftwareagentsdk} on GAIA~\citep{mialon2023gaiabenchmarkgeneralai}, PaperQA2~\citep{narayanan2024aviarytraininglanguageagents,lala2023paperqa} on LitQA2~\citep{skarlinski2024language}, LIFE-harness~\citep{xu2026adapting} on the airline and retail domains of $\tau^2$-bench~\citep{barres2025tau2benchevaluatingconversationalagents}, and the fixed agentfold harness from the JIT-Agent repository~\citep{zhang2026jitagentscalingharnessintelligence} on DeepPlanning-Shopping~\citep{zhang-etal-2026-deepplanning}. These settings cover general-purpose question answering, literature-based question answering, customer service, and shopping tasks. Table~\ref{tab:dataset-statistics} in Appendix~\ref{sec:implementation-details} summarizes the task counts and candidate module counts.

\paragraph{Data splits.} We select 50 development tasks, with stratified sampling when difficulty labels are available. The exception is the airline domain of $\tau^2$-bench, for which we use all 30 available training tasks. For validation, we use the full validation sets of both $\tau^2$-bench domains and all 38 LitQA2 tasks with accessible full-text articles, while sampling tasks stratified by difficulty for GAIA and DeepPlanning-Shopping. We follow official data splits where available. Otherwise, we construct disjoint development and validation sets from the task pool.

\paragraph{Evaluation protocol.} We use Qwen3.5-122B-A10B~\citep{qwen3.5} as the base model across all settings. After evaluating each single-module ablation once per development task, we use the saliency estimates to guide pruning and evaluate the resulting configurations on the validation set, with three trials per task by default. We evaluate a configuration after every three modules pruned, using a smaller first step when needed to reach the protected set exactly. When the protected set is nonempty, we additionally evaluate an all-pruned baseline that disables even the protected modules. Within each setting, model configuration and execution budgets remain fixed across configurations. Appendix~\ref{sec:implementation-details} provides implementation details.

\paragraph{Metrics.} We report task performance and token cost for each configuration. Tables report mean token cost across tasks and trials, whereas Figure \ref{fig:appendix-pruning-curves}-\ref{fig:main-pruning-tradeoffs} use the per-trial median across tasks to reduce the influence of extreme values. Task performance is measured by success rate on GAIA, LitQA2, and $\tau^2$-bench. For DeepPlanning-Shopping, we use the mean per-task match rate as the performance metric. Token cost includes input and output tokens from unsuccessful as well as successful attempts. Appendix~\ref{sec:implementation-details} describes the matching rules and token accounting.

\subsection{Main results}
\label{sec:experimental-results}

Figure~\ref{fig:appendix-pruning-curves} shows how task performance and token cost change as more modules are pruned. Across the five settings, many pruned configurations retain similar observed task performance, and several also reduce token cost relative to the full harness. For example, OpenHands achieves its highest success rate on GAIA with 24 of its 27 modules pruned, increasing from 29.44\% for the full harness to 35.00\%. JIT agentfold also retains similar performance across several smaller configurations with lower token cost, while LIFE-harness maintains much of its performance even when all candidate modules are pruned. These results suggest that substantial parts of an existing harness can be redundant for the tasks on which it is evaluated.

The extent of this redundancy differs across harnesses and task domains. PaperQA2 achieves its highest performance with the full harness, whereas the all-pruned configuration remains competitive in both LIFE-harness domains, with a modest performance decrease on airline and a slight increase on retail. The GAIA results further show that performance can improve under pruning but decline when too many modules are pruned. In this case, success rate reaches its highest value with 24 modules pruned, but falls from 35\% to 14.44\% when all 27 modules are pruned, despite the very low token cost. Token cost itself can also rise as more modules are removed. This occurs in PaperQA2's configuration with 15 modules pruned and JIT agentfold's all-pruned configuration, both of which are more expensive than their full harnesses.

\begin{figure}[!htb]
\centering
\begin{minipage}[t]{0.33\linewidth}
\vspace{0pt}
\centering
\includegraphics[width=\linewidth]{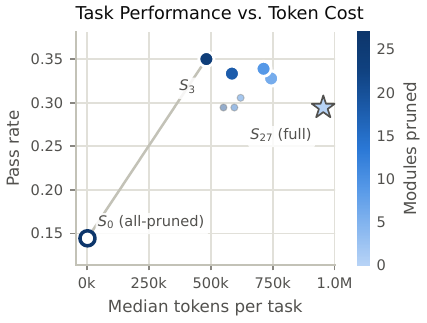}
\par {\footnotesize (a) OpenHands / GAIA}
\end{minipage}\hfill
\begin{minipage}[t]{0.33\linewidth}
\vspace{0pt}
\centering
\includegraphics[width=\linewidth]{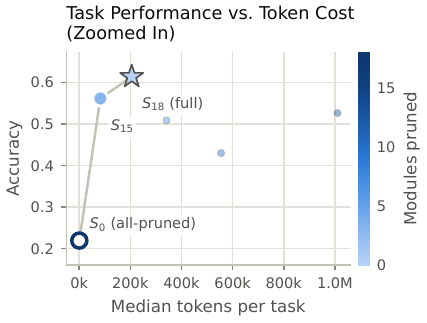}
\par {\footnotesize (b) PaperQA2 / LitQA2}
\end{minipage}\hfill
\begin{minipage}[t]{0.33\linewidth}
\vspace{0pt}
\centering
\includegraphics[width=\linewidth]{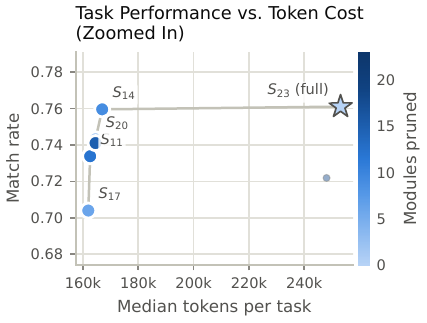}
\par {\footnotesize (c) JIT agentfold /\\DeepPlanning-Shopping}
\end{minipage}
\par\smallskip
\begin{minipage}[t]{0.33\linewidth}
\vspace{0pt}
\centering
\includegraphics[width=\linewidth]{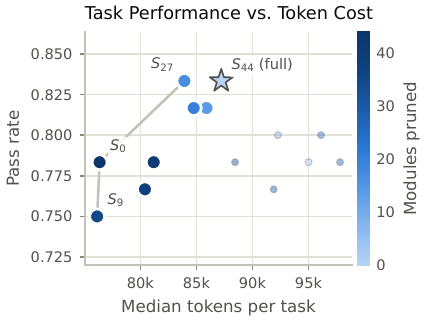}
\par {\footnotesize (d) LIFE-harness / $\tau^2$-bench airline}
\end{minipage}\hspace{0.02\linewidth}
\begin{minipage}[t]{0.33\linewidth}
\vspace{0pt}
\centering
\includegraphics[width=\linewidth]{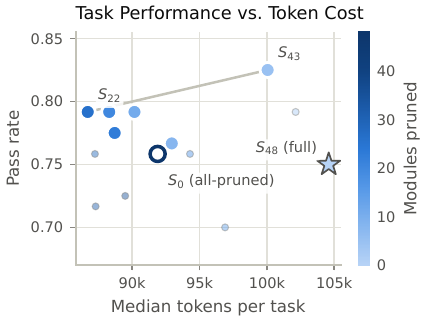}
\par {\footnotesize (e) LIFE-harness / $\tau^2$-bench retail}
\end{minipage}
\par\smallskip
\includegraphics[width=0.855\linewidth]{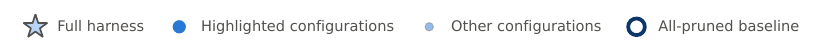}
\vspace{-5mm}
\caption{Task performance and token cost across pruned configurations. Stars mark full harnesses. Hollow circles mark the additional all-pruned baseline when the protected set is nonempty. Lines trace the Pareto frontier: no other evaluated configuration achieves higher task performance without increasing token cost, or lower token cost without reducing task performance. Panels (b) and (c) show zoomed-in views, omitting PaperQA2's $\mathcal{S}_3$ and JIT agentfold's $\mathcal{S}_0$, respectively, whose high token costs would compress the remaining configurations into a narrow region. Full-range plots are provided in Appendix Figure~\ref{fig:full-range-pruning-tradeoffs}.}
\label{fig:main-pruning-tradeoffs}
\vspace{-4mm}
\end{figure}

To examine task performance and token cost together, Figure~\ref{fig:main-pruning-tradeoffs} plots each configuration as a point, with token cost on the horizontal axis and task performance on the vertical axis. Configurations closer to the upper-left corner are more desirable because they achieve higher performance at lower cost. The lines trace the Pareto frontier, connecting configurations for which no other evaluated configuration improves one metric without worsening the other. The frontier reveals different performance--efficiency trade-offs across settings. For PaperQA2, configurations with lower token cost also have lower task performance than the full harness. In contrast, OpenHands and LIFE-harness retail include pruned configurations that achieve both higher observed performance and lower token cost, placing the full harness outside the frontier. The numerical results underlying these figures are reported in Appendix~\ref{sec:appendix-pruning-results}.

\begin{table}[!t]
\small
\centering
\setlength{\tabcolsep}{2.5pt}
\caption{Comparison of performance-oriented and efficiency-oriented pruning with one trial per validation task. $P$ is task performance (\%); $T$ is mean token cost ($10^3$). $\mathcal{S}'_k$ and $\mathcal{S}_k$ denote performance-oriented and efficiency-oriented configurations, respectively. Full and all-pruned results are shared between strategies. Bold values mark the best result in each metric column. Retail configurations are selected for brevity; complete results are provided in Appendix~\ref{sec:appendix-pruning-results}.}
\label{tab:pruning-strategies}
\begin{tabular*}{\linewidth}{@{\extracolsep{\fill}}lrr|lrr!{\vrule width 0.9pt}lrr|lrr@{}}
\toprule
\rowcolor[HTML]{D9E3F0}
\multicolumn{6}{c!{\vrule width 0.9pt}}{\textbf{JIT agentfold / DeepPlanning-Shopping}} & \multicolumn{6}{c}{\textbf{LIFE-harness / $\tau^2$-bench retail}} \\
\multicolumn{3}{c|}{Performance-oriented} & \multicolumn{3}{c!{\vrule width 0.9pt}}{Efficiency-oriented} & \multicolumn{3}{c|}{Performance-oriented} & \multicolumn{3}{c}{Efficiency-oriented} \\
\midrule
Config. & $P$ & $T$ & Config. & $P$ & $T$ & Config. & $P$ & $T$ & Config. & $P$ & $T$ \\
\midrule
$\mathcal{S}_{23}$\,{\scriptsize (full)} & 75.8 & 324 & $\mathcal{S}_{23}$\,{\scriptsize (full)} & 75.8 & 324 & $\mathcal{S}_{48}$\,{\scriptsize (full)} & 67.5 & 121 & $\mathcal{S}_{48}$\,{\scriptsize (full)} & 67.5 & 121 \\
$\mathcal{S}'_{22}$ & 77.7 & 278 & $\mathcal{S}_{20}$ & \textbf{76.2} & 197 & $\mathcal{S}'_{43}$ & 80.0 & 118 & $\mathcal{S}_{46}$ & 72.5 & 115 \\
$\mathcal{S}'_{19}$ & \textbf{82.5} & 305 & $\mathcal{S}_{17}$ & 67.2 & \textbf{196} & $\mathcal{S}'_{37}$ & 72.5 & 117 & $\mathcal{S}_{43}$ & \textbf{87.5} & 118 \\
$\mathcal{S}'_{16}$ & 76.9 & 219 & $\mathcal{S}_{14}$ & 71.9 & 198 & $\mathcal{S}'_{31}$ & 75.0 & 114 & $\mathcal{S}_{37}$ & 80.0 & 105 \\
$\mathcal{S}'_{13}$ & 72.7 & 223 & $\mathcal{S}_{11}$ & 70.0 & 206 & $\mathcal{S}'_{25}$ & 65.0 & 101 & $\mathcal{S}_{28}$ & 75.0 & 103 \\
$\mathcal{S}'_{10}$ & 70.7 & 265 & $\mathcal{S}_8$ & 74.2 & 209 & $\mathcal{S}'_{16}$ & 72.5 & 107 & $\mathcal{S}_{25}$ & 77.5 & \textbf{100} \\
$\mathcal{S}'_7$ & 68.5 & 246 & $\mathcal{S}_5$\,{\scriptsize (protected)} & 74.9 & 409 & $\mathcal{S}'_{10}$ & 75.0 & 99 & $\mathcal{S}_{19}$ & 82.5 & 107 \\
$\mathcal{S}'_4$\,{\scriptsize (protected)} & 63.3 & \textbf{146} &  &  &  & $\mathcal{S}'_7$\,{\scriptsize (protected)} & 70.0 & \textbf{96} & $\mathcal{S}_{13}$\,{\scriptsize (protected)} & 67.5 & 100 \\
$\mathcal{S}_0$\,{\scriptsize (all-pruned)} & 68.0 & 665 & $\mathcal{S}_0$\,{\scriptsize (all-pruned)} & 68.0 & 665 & $\mathcal{S}_0$\,{\scriptsize (all-pruned)} & \textbf{85.0} & 101 & $\mathcal{S}_0$\,{\scriptsize (all-pruned)} & 85.0 & 101 \\
\bottomrule
\end{tabular*}
\vspace{-4mm}
\end{table}

\subsection{Comparison of pruning strategies}
\label{sec:module-analysis}

We compare efficiency-oriented and performance-oriented pruning on JIT agentfold and LIFE-harness retail, using one trial per validation task (Table~\ref{tab:pruning-strategies}). The two strategies use the same saliency estimates but exchange their roles in defining the protected module set and determining the pruning order of unprotected modules. On DeepPlanning-Shopping, they identify different performance--efficiency trade-offs. Performance-oriented pruning reaches an observed match rate of 82.5\% at $\mathcal{S}'_{19}$, compared with 75.8\% for the full harness, while reducing mean token cost by 5.9\%. Efficiency-oriented pruning instead finds $\mathcal{S}_{20}$, which achieves a similar observed match rate of 76.2\% with a larger token reduction of 39.0\%.

On retail, however, efficiency-oriented pruning achieves the highest observed success rate at $\mathcal{S}_{43}$ (87.5\%). Performance-oriented pruning achieves the lowest token cost at $\mathcal{S}'_7$, reducing mean token cost by 21.2\% relative to the full harness with a success rate of 70.0\%. Among the configurations evaluated for performance-oriented pruning, the highest success rate is instead achieved by the all-pruned configuration (85.0\%). Together, these results show that the two strategies explore different performance--efficiency trade-offs, with their relative benefits depending on the harness and domain.

\subsection{Qualitative analysis}
\label{sec:pruning-cost}

We examine agent trajectories to understand how pruning changes the execution process and input context, and why these changes can either reduce or increase token cost.

\paragraph{When pruning improves efficiency.} Pruning can reduce both redundant actions and the amount of context processed at each step. In a DeepPlanning-Shopping task, JIT agentfold must select five clothing and footwear products satisfying the user's requirements at the lowest total price. The full harness includes a shopping cost strategy module that instructs the agent to search for cheaper alternatives. Even after assembling the correct cart, the agent makes 26 further searches and 26 product-detail queries without changing its selected products, eventually reaching the execution limit. With this instruction module removed, the agent selects the same five products at the same price and terminates after checking available discounts. Both runs achieve a match rate of 100\%, while model calls decrease from 43 to 14 and total token cost falls by 74\%. Here, additional optimization guidance leads to repeated checks that do not improve the final selection.

Unused capabilities can also add cost through their descriptions in the model's context. In a GAIA task, OpenHands reads a library inventory PDF and counts books by a specified author that are not on the shelves. Removing the version-control guidance, file editor, and skills (microagents) module preserves the same effective workflow. Both configurations use two terminal calls to obtain the PDF text and return the correct count after three model calls. The full harness nevertheless supplies descriptions of 64 available skills alongside its other instructions and tool definitions. Pruning these modules reduces the first model call's input from 19,706 to 12,953 tokens, with a similar reduction in subsequent calls. Total token cost decreases by 32\%, illustrating how a smaller context can improve efficiency even when the number of execution steps remains unchanged.

\begin{figure}[t]
\centering
\includegraphics[width=0.93\linewidth]{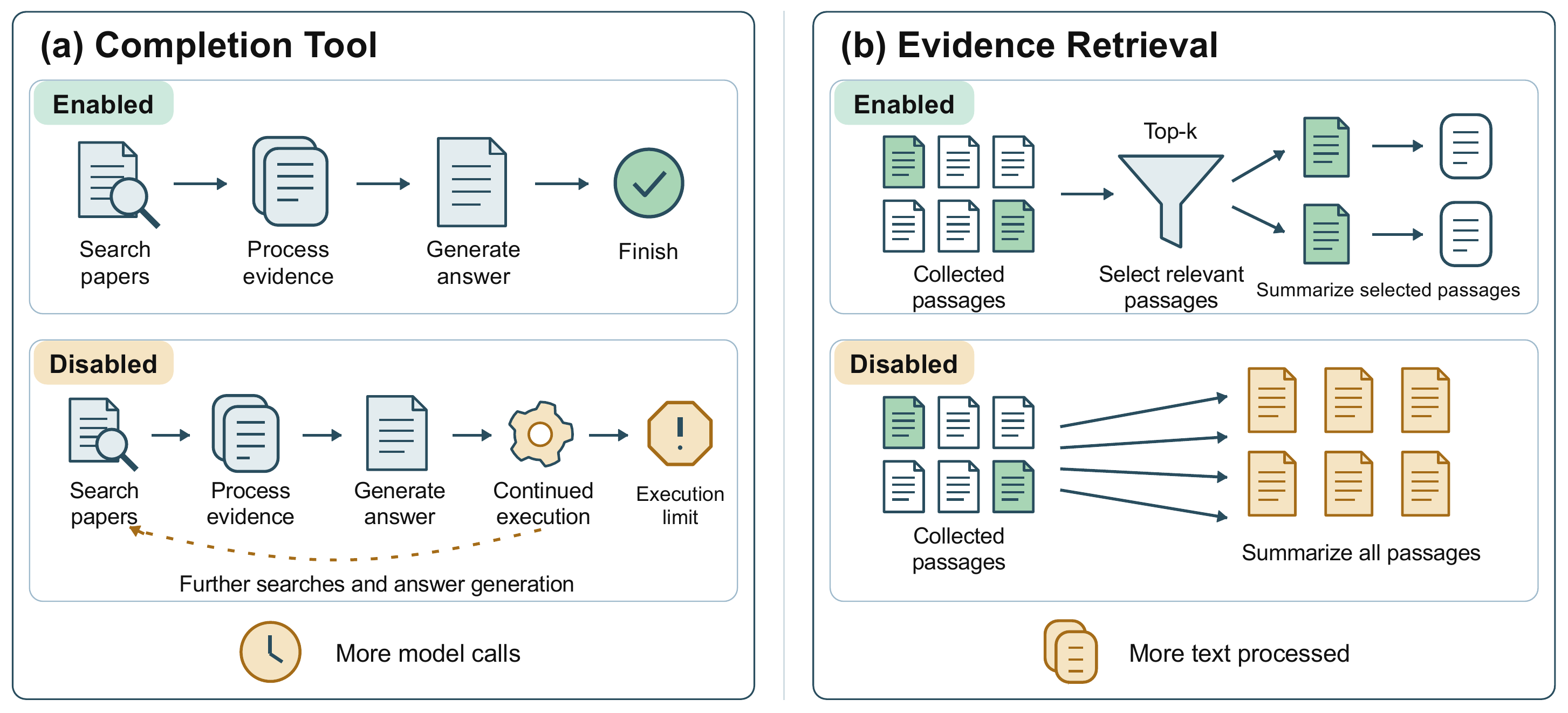}
\vspace{-2mm}
\caption{Examples from PaperQA2 showing how pruning can increase token cost. Disabling the completion tool prolongs execution after an answer is generated, while disabling evidence retrieval causes all collected passages to be summarized instead of a selected subset.}
\label{fig:paperqa2-cases}
\vspace{-5mm}
\end{figure}

\paragraph{When pruning increases cost.} PaperQA2 answers questions by searching scientific papers, extracting relevant information, and using it to generate an answer. Figure~\ref{fig:paperqa2-cases} illustrates two cases in which pruning increases the cost of this process. First, the completion tool allows the agent to submit its answer and end execution. When this tool is removed, the agent can still generate an answer but cannot terminate normally. In an inspected trajectory, it continues searching and processing papers and repeatedly generates answers until the step budget is exhausted.

Second, the evidence retrieval module selects relevant passages from the collected papers for summarization. When it is removed, all available passages are processed instead, including those that retrieval would have filtered out. Together, the two cases show how pruning can increase cost by extending execution after an answer has been generated or by increasing the amount of text processed during evidence summarization.

%% file: sections/conclusion.tex
\section{Conclusion}

We presented \algname, a saliency-based approach to analyzing redundancy in agent harnesses through pruning. Using single-module ablations to guide pruning, we found that smaller harness configurations can retain similar task performance while reducing token cost across diverse settings. The scope for pruning varies across harnesses and task domains, and removing more modules does not always improve efficiency. Our qualitative analysis further shows that pruning can reduce redundant actions and input context, but can also remove functionality that helps agents complete tasks efficiently. These findings suggest that harness design can benefit from examining which modules contribute to task performance and efficiency, alongside efforts to introduce new capabilities.

%% file: sections/acknowledgments.tex
\subsubsection*{Acknowledgments}
The UCSB team acknowledges support from the National Science Foundation (NSF) under Grant Nos. 2338252, 2302730, and 2619240.

%% file: sections/reproducibility.tex
\subsection*{Reproducibility statement}

Section~\ref{sec:methods} describes our pruning procedure, and Section~\ref{sec:experimental-setup} specifies the experimental setup and evaluation protocol. Appendix~\ref{sec:appendix-harness-modules} lists the candidate modules and their disabling behavior. Appendix~\ref{sec:implementation-details} provides implementation details, including data statistics, model settings, execution budgets, and token accounting. Appendix~\ref{sec:appendix-saliency-tests} details the statistical tests, while Appendices~\ref{sec:appendix-saliency-results} and~\ref{sec:appendix-pruning-results} report module saliency estimates and the evaluated pruning configurations with their results.

%% file: sections/ai_use.tex
\subsection*{AI use statement}

We used generative AI tools to assist with research ideation, implementation, and analysis, as well as drafting and refining the manuscript and preparing figures. This assistance included feedback on methodology and experiments, code development, and analysis of experimental results and agent trajectories. The authors reviewed and revised the AI-assisted content and take responsibility for the final text, methods, results, and conclusions.

%% file: sections/appendix.tex
\section{Appendix}

\subsection{Harness modules}
\label{sec:appendix-harness-modules}

We describe the candidate modules in each harness and the capabilities that remain available when all candidate modules are pruned. The Disabling notes column provides additional clarification where needed; a dash indicates that no additional note is provided.

\paragraph{All-pruned configurations.} Disabling all candidate modules leaves different capabilities available in different harnesses. LIFE-harness retains the native domain tools and policy, while JIT agentfold retains the shopping APIs and basic action protocol. In PaperQA2, all five agent tools are candidates, so the all-pruned configuration answers the question through a direct model call without tools.

\paragraph{PaperQA2.} Table~\ref{tab:paperqa2-modules} summarizes the 18 candidate modules used in our PaperQA2 experiments. Each row corresponds to one independently configurable module.

\begin{table}[htbp]
\small
\centering
\caption{Candidate modules in PaperQA2. Functions describe enabled behavior; disabling notes clarify alternative behavior where needed.}
\label{tab:paperqa2-modules}
\begin{tabular}{@{}>{\raggedright\arraybackslash}p{0.27\linewidth}>{\raggedright\arraybackslash}p{0.43\linewidth}>{\raggedright\arraybackslash}p{\dimexpr0.30\linewidth-4\tabcolsep\relax}@{}}
\toprule
Module & Function & Disabling notes \\
\midrule
Paper search tool & Searches for papers to add to the available document collection. & -- \\
Evidence gathering tool & Collects question-relevant evidence from the available papers. & -- \\
Answer generation tool & Synthesizes an answer using the collected evidence. & -- \\
Reset tool & Clears collected evidence so the agent can gather new evidence. & -- \\
Completion tool & Ends the agent loop using the latest proposed answer. & -- \\
\midrule
Agent system prompt & Supplies the agent's system-level instructions. & -- \\
Workflow guidance & Guides the agent's question-answering workflow. & Retains only the question and a minimal instruction to use tools. \\
Citation metadata context & Includes formatted citation metadata alongside each evidence passage. & -- \\
\midrule
Evidence retrieval & Selects relevant text through embedding-based retrieval. & Processes the available texts without this selection. \\
Evidence summarization & Produces LLM summaries of evidence passages. & Uses the passages directly. \\
Structured evidence output & Uses JSON-formatted evidence summaries. & Uses the plain-text summary path. \\
Automatic evidence gathering & Gathers evidence during answer generation if no evidence has been collected. & -- \\
Answer iteration & Includes the previous answer when generating a subsequent answer. & -- \\
Multimodal processing & Processes visual content in documents in addition to text. & -- \\
Document metadata extraction & Extracts document details such as titles and authors. & -- \\
\midrule
Increased evidence depth & Retrieves up to 20 evidence passages. & Uses the default limit of 10. \\
Increased concurrency & Allows up to 10 concurrent evidence-processing requests. & Uses the default concurrency of 4. \\
Larger text chunks & Uses 7,000-character text chunks. & Uses the default 5,000-character chunks. \\
\bottomrule
\end{tabular}
\end{table}

\paragraph{OpenHands.} Table~\ref{tab:openhands-modules} lists the 27 candidate modules in the OpenHands harness used for GAIA, including tools, instruction blocks, and context-management functionality.

\begingroup
\makeatletter
\def\LT@makecaption#1#2#3{%
  \LT@mcol\LT@cols c{\hbox to\z@{\hss\parbox[t]\LTcapwidth{%
    \reset@font\raggedright #1{#2: }#3\par\vskip\baselineskip}%
  \hss}}}
\makeatother
\setlength{\LTcapwidth}{\linewidth}
\small
\begin{longtable}{@{}>{\raggedright\arraybackslash}p{0.27\linewidth}>{\raggedright\arraybackslash}p{0.43\linewidth}>{\raggedright\arraybackslash}p{\dimexpr0.30\linewidth-4\tabcolsep\relax}@{}}
\caption{Candidate modules in OpenHands. Each row corresponds to one configurable module.}\label{tab:openhands-modules}\\
\toprule
Module & Function & Disabling notes \\
\midrule
\endfirsthead
\multicolumn{3}{l}{\textit{Table \thetable{} continued}} \\
\toprule
Module & Function & Disabling notes \\
\midrule
\endhead
\midrule
\multicolumn{3}{r}{\textit{Continued on next page}} \\
\endfoot
\bottomrule
\endlastfoot
Terminal tool & Executes shell commands in the workspace. & -- \\
Finish tool & Submits the final response and ends the agent loop. & -- \\
Think tool & Records a reasoning step without changing the environment. & -- \\
Browser tools & Navigate and interact with web pages, with accompanying usage guidance. & -- \\
Fetch tool & Retrieves web content from a URL. & -- \\
Search tool & Searches the web for relevant information. & -- \\
File editor & Views and edits local files. & -- \\
Task tracker & Maintains a structured list of tasks and their status. & -- \\
Delegation & Assigns subtasks to other agents. & -- \\
\midrule
Introductory prompt & Introduces the agent and its computer-interaction capabilities. & -- \\
Role guidance & Defines the agent's responsibilities and approach to user requests. & -- \\
Memory guidance & Encourages storing repository knowledge in a persistent file. & -- \\
Efficiency guidance & Encourages batching actions and using efficient search tools. & -- \\
File-system guidance & Guides file discovery, direct editing, and temporary-file handling. & -- \\
Code-quality guidance & Encourages focused changes and clear, maintainable code. & -- \\
Version-control guidance & Guides staging, committing, and safe Git operations. & -- \\
Pull-request guidance & Guides creating and updating pull requests. & -- \\
Problem-solving workflow & Organizes exploration, analysis, testing, implementation, and verification. & -- \\
Documentation guidance & Directs questions about OpenHands to its documentation. & -- \\
Security guidance & Supplies the configured security-policy instructions. & -- \\
Risk-assessment guidance & Instructs the agent to assess action security risks. & -- \\
External-service guidance & Guides API use and communication with external services. & -- \\
Environment-setup guidance & Guides installing applications and resolving missing dependencies. & -- \\
Troubleshooting guidance & Encourages revisiting possible causes when repeated attempts fail. & -- \\
Process-management guidance & Guides identifying and terminating specific processes. & -- \\
\midrule
History condensation & Summarizes conversation history to limit context growth. & Keeps conversation history without LLM summarization. \\
Skills (microagents) & Loads reusable instructions that provide additional task knowledge. & Skips loading the public skill library. \\
\end{longtable}
\endgroup

\paragraph{JIT agentfold.} Table~\ref{tab:jit-modules} lists the 23 candidate modules in the JIT agentfold harness used for DeepPlanning-Shopping. The benchmark's shopping APIs and final-answer interface remain available in every configuration.

\begingroup
\makeatletter
\def\LT@makecaption#1#2#3{%
  \LT@mcol\LT@cols c{\hbox to\z@{\hss\parbox[t]\LTcapwidth{%
    \reset@font\raggedright #1{#2: }#3\par\vskip\baselineskip}%
  \hss}}}
\makeatother
\setlength{\LTcapwidth}{\linewidth}
\small
\begin{longtable}{@{}>{\raggedright\arraybackslash}p{0.27\linewidth}>{\raggedright\arraybackslash}p{0.43\linewidth}>{\raggedright\arraybackslash}p{\dimexpr0.30\linewidth-4\tabcolsep\relax}@{}}
\caption{Candidate modules in JIT agentfold. The shopping guidance varies with task difficulty.}\label{tab:jit-modules}\\
\toprule
Module & Function & Disabling notes \\
\midrule
\endfirsthead
\multicolumn{3}{l}{\textit{Table \thetable{} continued}} \\
\toprule
Module & Function & Disabling notes \\
\midrule
\endhead
\midrule
\multicolumn{3}{r}{\textit{Continued on next page}} \\
\endfoot
\bottomrule
\endlastfoot
Initial planning & Generates a plan with goals and candidate solution paths. & Starts without an initial plan; the step budget is unchanged. \\
Periodic replanning & Reviews progress and updates the plan every eight steps. & -- \\
History folding & Replaces selected ranges of past steps with model-generated summaries. & Keeps the original history instead of applying requested summaries. \\
Forced final answer & Makes a final answer-generation call when the step limit is reached. & Returns an empty answer if the loop ends without a submitted answer. \\
Lenient JSON parsing & Repairs malformed action JSON. & Uses strict parsing. \\
\midrule
Task restatement & Repeats the original task in each step prompt. & -- \\
Compression guidance & Explains when and how to summarize past steps. & -- \\
Step-output example & Illustrates the expected action and compression format. & -- \\
\midrule
Role preamble & Introduces the agent's reasoning and tool-use cycle. & -- \\
Planning-model guidance & Explains goals, candidate paths, and updated plan summaries. & -- \\
Plan-execution guidance & Guides parallel goal execution and fallback between paths. & -- \\
Action guidance & Directs each step toward unresolved goals and suitable tool calls. & -- \\
Workflow examples & Illustrates tool use, history compression, and final answering. & -- \\
Result-processing guidance & Instructs the agent to combine tool outputs and verify goal completion. & -- \\
Final-answer guidance & Requires consolidating results after all goals are resolved. & -- \\
Tool-selection guidance & Encourages focused tool calls and reuse of existing observations. & -- \\
Tool-list summary & Lists tool names and descriptions in the system prompt. & The tool list in each step prompt remains available. \\
General action rules & Specifies argument use, avoidance of repeated calls, and answer language. & -- \\
\midrule
Code execution & Runs code in a workspace, with accompanying workspace instructions. & Also removes the accompanying workspace instructions. \\
\midrule
Shopping requirements & Guides identifying required products, quantities, and user constraints. & -- \\
Shopping cost strategy & Guides price minimization, budget handling, and coupon use according to task difficulty. & -- \\
Cart verification & Requires checking the cart rather than relying on intermediate calculations. & -- \\
Shopping output guidance & Specifies the product, cost, and coupon information to report. & -- \\
\end{longtable}
\endgroup

\paragraph{LIFE-harness.} Tables~\ref{tab:life-airline-modules} and~\ref{tab:life-retail-modules} list the 44 airline and 48 retail modules, respectively. Within each table, rows are grouped into pre-execution checks, tool-description hints, tool-result annotations and alerts, retrieved skills, and agent instructions. Each hint or annotation is a separate module from the underlying tool; the native domain tools and policy remain available when these modules are disabled.

\begingroup
\makeatletter
\def\LT@makecaption#1#2#3{%
  \LT@mcol\LT@cols c{\hbox to\z@{\hss\parbox[t]\LTcapwidth{%
    \reset@font\raggedright #1{#2: }#3\par\vskip\baselineskip}%
  \hss}}}
\makeatother
\setlength{\LTcapwidth}{\linewidth}
\small
\begin{longtable}{@{}>{\raggedright\arraybackslash}p{0.27\linewidth}>{\raggedright\arraybackslash}p{0.43\linewidth}>{\raggedright\arraybackslash}p{\dimexpr0.30\linewidth-4\tabcolsep\relax}@{}}
\caption{Candidate modules in LIFE-harness for the airline domain.}\label{tab:life-airline-modules}\\
\toprule
Module & Function & Disabling notes \\
\midrule
\endfirsthead
\multicolumn{3}{l}{\textit{Table \thetable{} continued}} \\
\toprule
Module & Function & Disabling notes \\
\midrule
\endhead
\midrule
\multicolumn{3}{r}{\textit{Continued on next page}} \\
\endfoot
\bottomrule
\endlastfoot
Booking payment limits & Checks the permitted numbers of certificates, credit cards, and gift cards. & -- \\
Booking itinerary check & Checks that flight legs form a connected itinerary with the required endpoints. & -- \\
Cancellation eligibility & Checks cancellation conditions and whether flights have already departed. & -- \\
Basic-economy changes & Blocks flight changes that violate basic-economy restrictions. & -- \\
Compensation eligibility & Checks membership, cabin, or insurance eligibility for compensation. & -- \\
Route preservation & Blocks changes to the reservation's origin or destination. & -- \\
Cabin-change eligibility & Blocks cabin changes after a flight has departed. & -- \\
Baggage reduction check & Prevents reducing the number of checked bags. & -- \\
Passenger limit & Enforces the five-passenger limit per reservation. & -- \\
Bookable flight status & Checks that requested flights have a bookable status. & -- \\
Booking baggage allowance & Checks the number of chargeable bags when booking. & -- \\
Booking payment total & Checks that payments match the fare, insurance, and baggage total. & -- \\
Flight-change payment type & Prevents using travel certificates to pay for flight changes. & -- \\
Flight-change payment capacity & Checks whether the payment method covers the net fare increase. & -- \\
Updated baggage allowance & Checks the number of chargeable bags when updating baggage. & -- \\
Passenger-count preservation & Prevents changing the number of passengers on an existing reservation. & -- \\
Compensable flight status & Checks for a delayed or cancelled flight that can justify compensation. & -- \\
Compensation amount check & Checks certificate amounts against compensation limits. & -- \\
Delay compensation prerequisite & Requires a completed change or cancellation before delay compensation. & -- \\
\midrule
Direct-flight search hint & Explains how to compare prices, schedules, and available seats. & -- \\
Connecting-flight search hint & Explains how to compare complete routes and check seats on both legs. & -- \\
Cancellation hint & Summarizes eligibility conditions and when human assistance is needed. & -- \\
Flight-update hint & Summarizes itinerary, cabin, and payment constraints for flight changes. & -- \\
Booking hint & Summarizes required booking details and payment, passenger, and baggage limits. & -- \\
Baggage-update hint & Explains requested bag additions and chargeable-bag counts. & -- \\
Passenger-update hint & Explains how to update passenger details without changing the count. & -- \\
Compensation hint & Summarizes compensation eligibility, timing, and amount limits. & -- \\
Transfer hint & Distinguishes requests requiring human assistance from supported operations. & -- \\
\midrule
User reservation summary & Adds a short summary of each reservation to user details. & -- \\
Reservation payment summary & Adds total charges and the payment-method breakdown to reservation details. & -- \\
Reservation policy context & Adds reservation-specific policy facts to tool results. & -- \\
Reservation flight summary & Highlights complete flight legs and reservation-wide cabin constraints. & -- \\
Cancellation refund summary & Reports the refund amount for each payment method after cancellation. & -- \\
Flight-update payment summary & Reports the net charge or refund after changing flights. & -- \\
Booking payment summary & Reports charges by payment method after booking. & -- \\
Repeated-failure alert & Warns the agent after repeated tool failures. & Tool failures are still returned to the agent. \\
\midrule
Exact-amount skill & Reminds the agent to communicate exact charges and refunds. & -- \\
Reservation-identification skill & Guides identifying the intended reservation before acting. & -- \\
Certificate-use skill & Guides certificate allocation and supported booking alternatives. & -- \\
Round-trip update skill & Guides submitting all flight legs together when changing a round trip. & -- \\
Cancellation-check skill & Guides checking cancellation conditions and completing confirmed changes. & -- \\
Budget-check skill & Guides calculating costs against the user's budget before making changes. & -- \\
Other-reservations skill & Guides checking all relevant reservations for broader itinerary requests. & -- \\
\midrule
Agent instructions & Provides general interaction instructions and organizes the domain-policy prompt. & Retains the domain policy and any retrieved task-specific guidance. \\
\end{longtable}
\endgroup

\begingroup
\makeatletter
\def\LT@makecaption#1#2#3{%
  \LT@mcol\LT@cols c{\hbox to\z@{\hss\parbox[t]\LTcapwidth{%
    \reset@font\raggedright #1{#2: }#3\par\vskip\baselineskip}%
  \hss}}}
\makeatother
\setlength{\LTcapwidth}{\linewidth}
\small
\begin{longtable}{@{}>{\raggedright\arraybackslash}p{0.27\linewidth}>{\raggedright\arraybackslash}p{0.43\linewidth}>{\raggedright\arraybackslash}p{\dimexpr0.30\linewidth-4\tabcolsep\relax}@{}}
\caption{Candidate modules in LIFE-harness for the retail domain.}\label{tab:life-retail-modules}\\
\toprule
Module & Function & Disabling notes \\
\midrule
\endfirsthead
\multicolumn{3}{l}{\textit{Table \thetable{} continued}} \\
\toprule
Module & Function & Disabling notes \\
\midrule
\endhead
\midrule
\multicolumn{3}{r}{\textit{Continued on next page}} \\
\endfoot
\bottomrule
\endlastfoot
Cancellation status check & Restricts cancellation to pending orders. & -- \\
Item-modification status check & Restricts item changes to orders still eligible for modification. & -- \\
Exchange status check & Restricts exchanges to eligible delivered orders. & -- \\
Return eligibility check & Checks delivered status and the permitted refund method. & -- \\
Cancellation reason check & Checks that the cancellation reason is an accepted value. & -- \\
Payment-change check & Requires the new payment method to differ from the current one. & -- \\
Payment balance check & Checks that a replacement gift card covers the full order total. & -- \\
Modification product-type check & Requires replacements to be variants of the same product. & -- \\
Exchange product-type check & Requires exchanged items to be variants of the same product. & -- \\
Modification item check & Checks that items selected for modification belong to the order. & -- \\
Exchange item check & Checks that items selected for exchange belong to the order. & -- \\
Return item check & Checks that items selected for return belong to the order. & -- \\
Distinct modification target & Rejects replacing an item with itself during modification. & -- \\
Distinct exchange target & Rejects exchanging an item for itself. & -- \\
Nonempty item list & Rejects empty lists of selected items. & Removes this check from modification, exchange, and return tools. \\
Nonempty replacement list & Rejects empty replacement-item lists. & Removes this check from both modification and exchange tools. \\
Item-list length check & Checks that selected and replacement lists have equal lengths. & Removes this check from both modification and exchange tools. \\
Replacement existence check & Checks replacement item identifiers against the product catalog. & Removes this check from both modification and exchange tools. \\
Item multiplicity check & Rejects duplicate selections exceeding the quantities in the order. & Removes this check from modification, exchange, and return tools. \\
\midrule
Cancellation hint & Explains whole-order cancellation, valid reasons, and the refund route. & -- \\
Item-modification hint & Summarizes replacement, availability, and payment constraints. & -- \\
Payment-change hint & Explains payment-method and gift-card balance requirements. & -- \\
Order-address hint & Explains eligible orders and how to copy the requested address. & -- \\
Exchange hint & Explains eligibility, variant selection, and grouping items in one exchange. & -- \\
Return hint & Summarizes return eligibility, item scope, and refund methods. & -- \\
Profile-address hint & Distinguishes profile changes from separate order-address changes. & -- \\
Transfer hint & Distinguishes unsupported requests from normal retail operations. & -- \\
\midrule
Order eligibility note & Flags statuses that restrict further modifications, exchanges, or returns. & -- \\
Order action guidance & Lists valid actions for pending and delivered orders. & -- \\
Exchange price difference & Highlights the price difference after a successful exchange. & -- \\
Return refund summary & Reports the refund amount after a successful return. & -- \\
Tracking information & Highlights tracking numbers in shipped-order details. & -- \\
Variant price extrema & Highlights the cheapest and most expensive available variants. & -- \\
Numeric-option extrema & Summarizes minimum and maximum numeric variant options. & -- \\
Available-variant count & Reports the number of available product variants. & -- \\
Item availability note & Clarifies that stock availability does not determine return eligibility. & -- \\
Cancellation refund summary & Reports the refund amount after a successful cancellation. & -- \\
Repeated-failure alert & Warns the agent after repeated tool failures. & Tool failures are still returned to the agent. \\
\midrule
Variant-counting skill & Guides counting all available variants separately from replacement preferences. & -- \\
Address-history skill & Guides locating and copying the requested address from existing records. & -- \\
Partial-cancellation skill & Explains when whole-order cancellation is an acceptable fallback. & -- \\
Bulk-request skill & Guides inspecting relevant orders and grouping requested changes by order. & -- \\
Multi-item exchange skill & Guides exchanging all requested items from one order in a single call. & -- \\
Numeric-option selection skill & Guides comparing numeric specifications while preserving other requirements. & -- \\
Category-scope skill & Keeps selected items within the category requested by the user. & -- \\
Pending-replacement skill & Distinguishes modifying an existing item from placing a new order. & -- \\
Confirmed-action skill & Reminds the agent to execute confirmed changes before claiming completion. & -- \\
\midrule
Agent instructions & Provides general interaction instructions and organizes the domain-policy prompt. & Retains the domain policy and any retrieved task-specific guidance. \\
\end{longtable}
\endgroup

\subsection{Implementation details}
\label{sec:implementation-details}

We describe the inference setup, harness implementations, and evaluation metrics used in our experiments. Table~\ref{tab:dataset-statistics} summarizes the task counts and candidate module counts. For the retail domain of $\tau^2$-bench, two of the 50 sampled development tasks are excluded because their evaluation outputs could not be parsed, leaving 48 evaluable tasks.

\begin{table}[!htb]
\small
\centering
\caption{Dataset and harness statistics. Development and validation denote the numbers of tasks in the respective sets, and $m$ is the number of candidate modules.}
\label{tab:dataset-statistics}
\setlength{\tabcolsep}{4pt}
\begin{tabular*}{\linewidth}{@{\extracolsep{\fill}}llrrr@{}}
\toprule
Benchmark & Harness & Development & Validation & $m$ \\
\midrule
GAIA & OpenHands & 50 & 60 & 27 \\
LitQA2 & PaperQA2 & 50 & 38 & 18 \\
$\tau^2$-bench airline & LIFE-harness & 30 & 20 & 44 \\
$\tau^2$-bench retail & LIFE-harness & 48 & 40 & 48 \\
DeepPlanning-Shopping & JIT agentfold & 50 & 30 & 23 \\
\bottomrule
\end{tabular*}
\end{table}

\paragraph{Inference setup.} We serve the FP8 checkpoint of Qwen3.5-122B-A10B~\citep{qwen3.5} through vLLM~\citep{kwon2023efficient}, with thinking disabled. Most experiments use NVIDIA H100 GPUs with tensor parallelism across two or four GPUs; some runs use A100 GPUs. Table~\ref{tab:inference-settings} summarizes the main settings. OpenHands uses temperature 0.7 and top-$p$ 0.8, while the other harnesses use temperature 0. Single-module ablations use one trial per task, main results use three trials, and the pruning-strategy comparison uses one trial. The two strategies share the same full and all-pruned results in Table~\ref{tab:pruning-strategies}.

\begin{table}[htbp]
\small
\centering
\setlength{\tabcolsep}{4pt}
\caption{Inference settings and execution budgets. Output limits are per model call. A dash denotes no explicit client-side output-token limit. Step counts follow each harness's execution loop.}
\label{tab:inference-settings}
\begin{tabular*}{\linewidth}{@{\extracolsep{\fill}}lrrl@{}}
\toprule
\multicolumn{4}{l}{Model: Qwen3.5-122B-A10B-FP8; inference engine: vLLM; thinking: disabled} \\
\multicolumn{4}{l}{Typical deployment: $2$ or $4\times$ NVIDIA H100 GPUs with tensor parallelism} \\
\midrule
Harness / benchmark & Temperature & Max output tokens & Execution budget \\
\midrule
OpenHands / GAIA & 0.7 & -- & 30 iterations \\
PaperQA2 / LitQA2 & 0 & 4,096 & 500 seconds \\
LIFE-harness / Airline & 0 & 16,384 & 200 steps \\
LIFE-harness / Retail & 0 & 16,384 & 200 steps \\
JIT agentfold / Shopping & 0 & 32,000 & 40 steps \\
\bottomrule
\end{tabular*}
\end{table}

PaperQA2 uses a server context limit of 65,536 tokens; LIFE-harness and JIT agentfold use 262,144 tokens. OpenHands leaves client-side input and output token limits unset. JIT additionally limits model calls to three times its step budget, while LIFE-harness permits at most ten execution errors.

\paragraph{Harness implementations.} For OpenHands, we use the GAIA evaluation pipeline provided in the OpenHands benchmarks repository, with local task workspaces and SearXNG~\citep{searxng} as the search backend. We apply module switches within this pipeline and use its answer extraction and scoring procedure. For JIT, we use the fixed \texttt{agentfold} seed harness provided by the JIT-Agent repository, which includes planning, tool execution, and history compression. We evaluate this harness and its pruned variants directly, without running the repository's harness optimization process. The official DeepPlanning-Shopping APIs and evaluator are used throughout.

PaperQA2 uses its \texttt{high\_quality} preset with a fixed local corpus and prebuilt document index. Its agent, answer-generation, summarization, and metadata-processing model calls use the same Qwen model. Retrieval uses \texttt{multi-qa-MiniLM-L6-cos-v1} embeddings. LIFE-harness uses the native $\tau^2$-bench task environments and domain policies, with its additional checks, tool descriptions, annotations, and skills controlled by the module switches. Its evaluation includes the benchmark's natural-language assertion checks; the user simulator and assertion judge use the same Qwen model with thinking disabled.

\paragraph{Token accounting.} Table~\ref{tab:token-accounting} specifies the model calls included in reported token cost. We sum input and output tokens across the recorded calls for each task, including unsuccessful attempts. These counts measure token usage rather than elapsed execution time.

\begin{table}[htbp]
\small
\centering
\caption{Token accounting for each harness.}
\label{tab:token-accounting}
\begin{tabular}{@{}p{0.23\linewidth}p{0.73\linewidth}@{}}
\toprule
Harness & Included token usage \\
\midrule
OpenHands & Model input and output tokens recorded by the SDK during task execution, including recorded delegated-agent calls. \\
PaperQA2 & Model input and output tokens in the question-answering session's token counts. Offline corpus preprocessing and index construction are excluded. \\
LIFE-harness & Agent and user-simulator model calls. The separate evaluation judge is excluded. \\
JIT agentfold & All model calls made by the harness during the task, including planning, action generation, and history compression when enabled. \\
\bottomrule
\end{tabular}
\end{table}

\paragraph{Shopping metric.} For DeepPlanning-Shopping, we use the match rate returned by the JIT repository's evaluator. For each task, the numerator is the number of matched distinct product IDs plus the number of matched coupon entries, and the denominator is the number of distinct product IDs plus the number of coupon entries in the reference answer. A coupon entry matches when both its name and quantity match the reference. Extra product IDs in the cart are recorded separately and do not reduce this score. We average the per-task match rates across evaluated tasks and trials.

\subsection{Statistical tests for module saliency}
\label{sec:appendix-saliency-tests}

We compute performance and efficiency saliency using two one-sided paired t-tests on the development tasks. Each single-module ablation outcome is paired with the median reference on the same task, giving the differences defined in Equation~\ref{eq:paired-differences}. The task is the unit of analysis, and each test is implemented as a one-sample t-test on these paired differences.

For $a\in\{\mathrm{perf},\mathrm{tok}\}$, let $\Delta_i^a=\mathbb{E}_t[d_{ti}^a]$ denote the population mean difference for module $i$. The performance test assesses whether disabling the module yields lower task scores than the reference:
\begin{equation}
H_{0,i}^{\mathrm{perf}}:\Delta_i^{\mathrm{perf}}\geq 0,
\qquad
H_{1,i}^{\mathrm{perf}}:\Delta_i^{\mathrm{perf}}<0.
\label{eq:appendix-perf-hypotheses}
\end{equation}
The efficiency test assesses whether disabling the module yields higher token cost than the reference:
\begin{equation}
H_{0,i}^{\mathrm{eff}}:\Delta_i^{\mathrm{tok}}\leq 0,
\qquad
H_{1,i}^{\mathrm{eff}}:\Delta_i^{\mathrm{tok}}>0.
\label{eq:appendix-eff-hypotheses}
\end{equation}

For each test, we estimate the mean and variance of the paired differences across the $n$ development tasks as
\begin{equation}
\widehat{\Delta}_i^a=\frac{1}{n}\sum_{t=1}^{n}d_{ti}^a,
\qquad
(s_i^a)^2=\frac{1}{n-1}\sum_{t=1}^{n}\bigl(d_{ti}^a-\widehat{\Delta}_i^a\bigr)^2.
\label{eq:appendix-paired-moments}
\end{equation}
For $s_i^a>0$, the estimated standard error of the mean difference is $s_i^a/\sqrt{n}$, and the test statistic is
\begin{equation}
T_i^a=\frac{\widehat{\Delta}_i^a}{s_i^a/\sqrt{n}}.
\label{eq:appendix-paired-statistic}
\end{equation}
Using the Student t reference distribution with $n-1$ degrees of freedom, whose cumulative distribution function is $F_{n-1}$, we compute the one-sided p-values as
\begin{equation}
p_i^{\mathrm{perf}}=F_{n-1}(T_i^{\mathrm{perf}}),
\qquad
p_i^{\mathrm{eff}}=1-F_{n-1}(T_i^{\mathrm{tok}}).
\label{eq:appendix-saliency-pvalues}
\end{equation}

\paragraph{Degenerate paired differences.} When the task-paired differences for a module are constant across tasks, their sample standard deviation is zero and the usual t-statistic is undefined. We handle this case using the following computational convention: if all differences are zero, we assign $p=1$; otherwise, we assign $p=0$ if the common difference is negative for the performance test or positive for the efficiency test, and $p=1$ if it has the opposite sign.

\subsection{Module saliency results}
\label{sec:appendix-saliency-results}

We report task performance and token cost for individual module ablations on the development tasks, together with the p-values used to estimate performance and efficiency saliency.

\begingroup
\setlength{\LTcapwidth}{\linewidth}
\small
\setlength{\tabcolsep}{4pt}
\begin{longtable}{@{}lrrrr@{}}
\caption{Single-module ablation results for JIT agentfold on 50 DeepPlanning-Shopping development tasks, with one trial per task. $P$ is mean match rate (\%); $T$ is mean token cost ($10^3$) after disabling the named module. The one-sided paired t-test p-values $p^{\mathrm{perf}}$ and $p^{\mathrm{eff}}$ quantify performance and efficiency saliency relative to the per-task median reference. Bold p-values satisfy $p\leq 0.05$.}\label{tab:jit-single-module-results}\\
\toprule
Module disabled & $P$ & $T$ & $p^{\mathrm{perf}}$ & $p^{\mathrm{eff}}$ \\
\midrule
\endfirsthead
\multicolumn{5}{l}{\textit{Table \thetable{} continued}} \\
\toprule
Module disabled & $P$ & $T$ & $p^{\mathrm{perf}}$ & $p^{\mathrm{eff}}$ \\
\midrule
\endhead
\midrule
\multicolumn{5}{r}{\textit{Continued on next page}} \\
\endfoot
\bottomrule
\endlastfoot
Initial planning & 73.70 & 328.7 & 0.737 & 0.539 \\
Periodic replanning & 67.97 & 367.0 & \textbf{0.028} & 0.075 \\
History folding & 76.40 & 535.1 & 0.923 & $\mathbf{<0.001}$ \\
Forced final answer & 71.13 & 359.2 & 0.314 & 0.058 \\
Lenient JSON parsing & 71.50 & 325.6 & 0.344 & 0.599 \\
Task restatement & 67.23 & 315.8 & \textbf{0.023} & 0.832 \\
Compression guidance & 71.33 & 344.7 & 0.347 & 0.243 \\
Step-output example & 50.80 & 787.0 & $\mathbf{<0.001}$ & $\mathbf{<0.001}$ \\
Role preamble & 73.33 & 369.0 & 0.620 & 0.061 \\
Planning-model guidance & 69.03 & 367.7 & 0.126 & 0.062 \\
Plan-execution guidance & 73.20 & 329.7 & 0.599 & 0.518 \\
Action guidance & 65.67 & 373.6 & 0.063 & \textbf{0.028} \\
Workflow examples & 61.47 & 371.8 & \textbf{0.002} & \textbf{0.021} \\
Result-processing guidance & 73.40 & 342.7 & 0.628 & 0.293 \\
Final-answer guidance & 70.53 & 323.2 & 0.274 & 0.631 \\
Tool-selection guidance & 69.87 & 324.7 & 0.121 & 0.619 \\
Tool-list summary & 66.73 & 331.5 & 0.059 & 0.484 \\
General action rules & 68.33 & 352.5 & 0.092 & 0.120 \\
Code execution & 71.33 & 277.5 & 0.378 & 0.998 \\
Shopping requirements & 69.50 & 310.1 & 0.174 & 0.818 \\
Shopping cost strategy & 72.50 & 285.1 & 0.507 & 0.990 \\
Cart verification & 66.03 & 320.0 & \textbf{0.033} & 0.688 \\
Shopping output guidance & 70.83 & 348.4 & 0.286 & 0.211 \\
\end{longtable}
\endgroup

\begingroup
\setlength{\LTcapwidth}{\linewidth}
\small
\setlength{\tabcolsep}{4pt}
\begin{longtable}{@{}lrrrr@{}}
\caption{Single-module ablation results for PaperQA2 on 50 LitQA2 development tasks, with one trial per task. $P$ is success rate (\%); $T$ is mean token cost ($10^3$) after disabling the named module. The one-sided paired t-test p-values $p^{\mathrm{perf}}$ and $p^{\mathrm{eff}}$ quantify performance and efficiency saliency. Both tests use the per-task median reference. Bold p-values satisfy $p\leq 0.05$.}\label{tab:paperqa2-single-module-results}\\
\toprule
Module disabled & $P$ & $T$ & $p^{\mathrm{perf}}$ & $p^{\mathrm{eff}}$ \\
\midrule
\endfirsthead
\multicolumn{5}{l}{\textit{Table \thetable{} continued}} \\
\toprule
Module disabled & $P$ & $T$ & $p^{\mathrm{perf}}$ & $p^{\mathrm{eff}}$ \\
\midrule
\endhead
\midrule
\multicolumn{5}{r}{\textit{Continued on next page}} \\
\endfoot
\bottomrule
\endlastfoot
Paper search tool & 0.00 & 1318.1 & $\mathbf{<0.001}$ & $\mathbf{<0.001}$ \\
Evidence gathering tool & 38.00 & 1466.1 & 0.664 & $\mathbf{<0.001}$ \\
Answer generation tool & 0.00 & 1232.2 & $\mathbf{<0.001}$ & $\mathbf{<0.001}$ \\
Reset tool & 42.00 & 184.0 & 0.887 & 0.215 \\
Completion tool & 80.00 & 997.9 & 1.000 & $\mathbf{<0.001}$ \\
Agent system prompt & 42.00 & 311.3 & 0.936 & 0.072 \\
Workflow guidance & 38.00 & 1206.9 & 0.741 & \textbf{0.001} \\
Citation metadata context & 32.00 & 138.5 & 0.291 & 0.487 \\
Evidence retrieval & 42.00 & 541.1 & 0.936 & \textbf{0.011} \\
Evidence summarization & 12.00 & 538.3 & $\mathbf{<0.001}$ & \textbf{0.044} \\
Structured evidence output & 36.00 & 200.1 & 0.594 & 0.248 \\
Automatic evidence gathering & 32.00 & 184.9 & 0.205 & 0.189 \\
Answer iteration & 34.00 & 214.9 & 0.406 & 0.256 \\
Multimodal processing & 32.00 & 142.7 & 0.313 & 0.440 \\
Document metadata extraction & 36.00 & 293.1 & 0.564 & 0.081 \\
Increased evidence depth & 40.00 & 80.7 & 0.952 & 0.861 \\
Increased concurrency & 40.00 & 260.8 & 0.952 & 0.114 \\
Larger text chunks & 38.00 & 324.4 & 0.741 & 0.140 \\
\end{longtable}
\endgroup

\begingroup
\setlength{\LTcapwidth}{\linewidth}
\small
\setlength{\tabcolsep}{4pt}
\begin{longtable}{@{}lrrrr@{}}
\caption{Single-module ablation results for OpenHands on 50 GAIA development tasks, with one trial per task. $P$ is success rate (\%); $T$ is mean token cost ($10^3$) after disabling the named module. The one-sided paired t-test p-values $p^{\mathrm{perf}}$ and $p^{\mathrm{eff}}$ quantify performance and efficiency saliency. Both tests use the per-task median reference. Bold p-values satisfy $p\leq 0.05$.}\label{tab:gaia-single-module-results}\\
\toprule
Module disabled & $P$ & $T$ & $p^{\mathrm{perf}}$ & $p^{\mathrm{eff}}$ \\
\midrule
\endfirsthead
\multicolumn{5}{l}{\textit{Table \thetable{} continued}} \\
\toprule
Module disabled & $P$ & $T$ & $p^{\mathrm{perf}}$ & $p^{\mathrm{eff}}$ \\
\midrule
\endhead
\midrule
\multicolumn{5}{r}{\textit{Continued on next page}} \\
\endfoot
\bottomrule
\endlastfoot
Terminal tool & 50.00 & 576.6 & 0.084 & \textbf{0.005} \\
Finish tool & 58.00 & 470.1 & 0.371 & 0.481 \\
Think tool & 60.00 & 483.0 & 0.500 & 0.330 \\
Browser tools & 62.00 & 480.4 & 0.670 & 0.347 \\
Fetch tool & 54.00 & 514.8 & \textbf{0.042} & 0.069 \\
Search tool & 44.00 & 639.9 & \textbf{0.022} & \textbf{0.003} \\
File editor & 70.00 & 437.1 & 0.971 & 0.917 \\
Task tracker & 56.00 & 515.7 & 0.161 & 0.089 \\
Delegation & 52.00 & 490.5 & \textbf{0.022} & 0.195 \\
Introductory prompt & 62.00 & 471.7 & 0.839 & 0.448 \\
Role guidance & 52.00 & 524.0 & 0.052 & \textbf{0.050} \\
Memory guidance & 60.00 & 444.0 & 0.500 & 0.797 \\
Efficiency guidance & 68.00 & 508.8 & 0.920 & 0.114 \\
File-system guidance & 56.00 & 521.3 & 0.210 & 0.080 \\
Code-quality guidance & 62.00 & 475.1 & 0.670 & 0.402 \\
Version-control guidance & 68.00 & 431.5 & 0.948 & 0.895 \\
Pull-request guidance & 66.00 & 479.2 & 0.958 & 0.370 \\
Problem-solving workflow & 60.00 & 500.9 & 0.500 & 0.140 \\
Documentation guidance & 58.00 & 444.3 & 0.355 & 0.811 \\
Security guidance & 62.00 & 493.7 & 0.670 & 0.162 \\
Risk-assessment guidance & 66.00 & 456.1 & 0.909 & 0.628 \\
External-service guidance & 64.00 & 462.3 & 0.790 & 0.604 \\
Environment-setup guidance & 58.00 & 487.0 & 0.330 & 0.322 \\
Troubleshooting guidance & 56.00 & 488.6 & 0.161 & 0.263 \\
Process-management guidance & 60.00 & 530.7 & 0.500 & 0.066 \\
History condensation & 62.00 & 525.5 & 0.670 & 0.106 \\
Skills (microagents) & 66.00 & 392.3 & 0.958 & 0.991 \\
\end{longtable}
\endgroup

\begingroup
\setlength{\LTcapwidth}{\linewidth}
\small
\setlength{\tabcolsep}{4pt}
\begin{longtable}{@{}lrrrr@{}}
\caption{Single-module ablation results for LIFE-harness on $\tau^2$-bench airline on 30 development tasks, with one trial per task. $P$ is success rate (\%); $T$ is mean token cost ($10^3$) after disabling the named module. The one-sided paired t-test p-values $p^{\mathrm{perf}}$ and $p^{\mathrm{eff}}$ quantify performance and efficiency saliency. Both tests use the per-task median reference. Bold p-values satisfy $p\leq 0.05$.}\label{tab:airline-single-module-results}\\
\toprule
Module disabled & $P$ & $T$ & $p^{\mathrm{perf}}$ & $p^{\mathrm{eff}}$ \\
\midrule
\endfirsthead
\multicolumn{5}{l}{\textit{Table \thetable{} continued}} \\
\toprule
Module disabled & $P$ & $T$ & $p^{\mathrm{perf}}$ & $p^{\mathrm{eff}}$ \\
\midrule
\endhead
\midrule
\multicolumn{5}{r}{\textit{Continued on next page}} \\
\endfoot
\bottomrule
\endlastfoot
Booking payment limits & 86.67 & 131.6 & 0.837 & \textbf{0.024} \\
Booking itinerary check & 83.33 & 112.4 & 1.000 & 0.760 \\
Cancellation eligibility & 83.33 & 117.3 & 0.500 & 0.501 \\
Basic-economy changes & 86.67 & 120.6 & 0.837 & 0.267 \\
Compensation eligibility & 83.33 & 120.2 & 0.500 & 0.375 \\
Route preservation & 83.33 & 121.1 & 1.000 & 0.319 \\
Cabin-change eligibility & 80.00 & 121.5 & 0.163 & 0.229 \\
Baggage reduction check & 86.67 & 120.8 & 0.837 & 0.207 \\
Passenger limit & 80.00 & 124.3 & 0.163 & 0.196 \\
Bookable flight status & 86.67 & 126.0 & 0.714 & 0.065 \\
Booking baggage allowance & 93.33 & 122.0 & 0.958 & 0.191 \\
Booking payment total & 86.67 & 118.5 & 0.714 & 0.436 \\
Flight-change payment type & 83.33 & 122.8 & 1.000 & 0.213 \\
Flight-change payment capacity & 83.33 & 126.1 & 0.500 & 0.115 \\
Updated baggage allowance & 76.67 & 116.4 & 0.080 & 0.583 \\
Passenger-count preservation & 86.67 & 138.7 & 0.714 & 0.117 \\
Compensable flight status & 90.00 & 145.7 & 0.920 & 0.129 \\
Compensation amount check & 86.67 & 118.1 & 0.837 & 0.440 \\
Delay compensation prerequisite & 80.00 & 119.7 & 0.163 & 0.285 \\
Direct-flight search hint & 80.00 & 117.8 & 0.286 & 0.471 \\
Connecting-flight search hint & 86.67 & 121.8 & 0.837 & 0.231 \\
Cancellation hint & 76.67 & 124.7 & 0.080 & 0.116 \\
Flight-update hint & 80.00 & 127.6 & 0.163 & 0.087 \\
Booking hint & 83.33 & 114.2 & 0.500 & 0.661 \\
Baggage-update hint & 80.00 & 123.1 & 0.163 & 0.163 \\
Passenger-update hint & 80.00 & 121.4 & 0.163 & 0.282 \\
Compensation hint & 83.33 & 136.7 & 0.500 & 0.171 \\
Transfer hint & 86.67 & 139.2 & 0.837 & 0.143 \\
User reservation summary & 86.67 & 123.2 & 0.714 & 0.214 \\
Reservation payment summary & 80.00 & 120.5 & 0.286 & 0.328 \\
Reservation policy context & 76.67 & 128.9 & 0.080 & 0.051 \\
Reservation flight summary & 86.67 & 110.1 & 0.837 & 0.901 \\
Cancellation refund summary & 86.67 & 117.4 & 0.837 & 0.496 \\
Flight-update payment summary & 80.00 & 147.6 & 0.163 & 0.084 \\
Booking payment summary & 86.67 & 121.6 & 0.837 & 0.208 \\
Repeated-failure alert & 86.67 & 131.2 & 0.714 & \textbf{0.045} \\
Exact-amount skill & 83.33 & 124.6 & 0.500 & 0.234 \\
Reservation-identification skill & 86.67 & 125.2 & 0.837 & 0.212 \\
Certificate-use skill & 80.00 & 127.1 & 0.163 & \textbf{0.047} \\
Round-trip update skill & 83.33 & 127.0 & 1.000 & \textbf{0.044} \\
Cancellation-check skill & 86.67 & 118.1 & 0.837 & 0.435 \\
Budget-check skill & 83.33 & 125.6 & 0.500 & 0.104 \\
Other-reservations skill & 83.33 & 116.9 & 0.500 & 0.523 \\
Agent instructions & 83.33 & 110.1 & 1.000 & 0.888 \\
\end{longtable}
\endgroup

\begingroup
\setlength{\LTcapwidth}{\linewidth}
\small
\setlength{\tabcolsep}{4pt}
\begin{longtable}{@{}lrrrr@{}}
\caption{Single-module ablation results for LIFE-harness on the $\tau^2$-bench retail development set, with one trial per task. $P$ is success rate (\%); $T$ is mean token cost ($10^3$) after disabling the named module. The one-sided paired t-test p-values $p^{\mathrm{perf}}$ and $p^{\mathrm{eff}}$ quantify performance and efficiency saliency. Both tests use the per-task median reference. Bold p-values satisfy $p\leq 0.05$.}\label{tab:retail-single-module-results}\\
\toprule
Module disabled & $P$ & $T$ & $p^{\mathrm{perf}}$ & $p^{\mathrm{eff}}$ \\
\midrule
\endfirsthead
\multicolumn{5}{l}{\textit{Table \thetable{} continued}} \\
\toprule
Module disabled & $P$ & $T$ & $p^{\mathrm{perf}}$ & $p^{\mathrm{eff}}$ \\
\midrule
\endhead
\midrule
\multicolumn{5}{r}{\textit{Continued on next page}} \\
\endfoot
\bottomrule
\endlastfoot
Cancellation status check & 81.25 & 124.0 & 0.210 & 0.304 \\
Item-modification status check & 79.17 & 122.0 & 0.091 & 0.551 \\
Exchange status check & 77.08 & 123.0 & 0.052 & 0.413 \\
Return eligibility check & 72.92 & 123.9 & \textbf{0.016} & 0.274 \\
Cancellation reason check & 79.17 & 127.0 & 0.131 & 0.098 \\
Payment-change check & 77.08 & 129.5 & \textbf{0.022} & 0.095 \\
Payment balance check & 85.42 & 122.1 & 0.500 & 0.534 \\
Modification product-type check & 81.25 & 122.9 & 0.210 & 0.402 \\
Exchange product-type check & 75.00 & 123.4 & \textbf{0.012} & 0.317 \\
Modification item check & 70.83 & 121.2 & \textbf{0.009} & 0.689 \\
Exchange item check & 77.08 & 130.1 & 0.080 & \textbf{0.011} \\
Return item check & 81.25 & 121.9 & 0.210 & 0.558 \\
Distinct modification target & 75.00 & 121.7 & \textbf{0.012} & 0.611 \\
Distinct exchange target & 81.25 & 126.1 & 0.161 & 0.052 \\
Nonempty item list & 77.08 & 125.2 & 0.052 & 0.112 \\
Nonempty replacement list & 85.42 & 122.2 & 0.500 & 0.521 \\
Item-list length check & 83.33 & 123.9 & 0.355 & 0.221 \\
Replacement existence check & 85.42 & 121.4 & 0.500 & 0.677 \\
Item multiplicity check & 81.25 & 132.0 & 0.161 & \textbf{0.031} \\
Cancellation hint & 79.17 & 121.2 & 0.091 & 0.685 \\
Item-modification hint & 83.33 & 120.3 & 0.355 & 0.834 \\
Payment-change hint & 87.50 & 122.6 & 0.670 & 0.457 \\
Order-address hint & 79.17 & 118.2 & 0.161 & 0.872 \\
Exchange hint & 75.00 & 125.5 & \textbf{0.048} & 0.214 \\
Return hint & 83.33 & 119.0 & 0.355 & 0.926 \\
Profile-address hint & 79.17 & 122.4 & \textbf{0.042} & 0.486 \\
Transfer hint & 81.25 & 122.6 & 0.210 & 0.457 \\
Order eligibility note & 77.08 & 126.9 & 0.052 & 0.140 \\
Order action guidance & 79.17 & 128.9 & 0.161 & \textbf{0.023} \\
Exchange price difference & 70.83 & 125.3 & \textbf{0.017} & 0.123 \\
Return refund summary & 77.08 & 124.0 & 0.052 & 0.269 \\
Tracking information & 81.25 & 123.8 & 0.210 & 0.281 \\
Variant price extrema & 68.75 & 130.1 & \textbf{0.005} & \textbf{0.036} \\
Numeric-option extrema & 83.33 & 123.7 & 0.355 & 0.289 \\
Available-variant count & 75.00 & 122.3 & \textbf{0.048} & 0.499 \\
Item availability note & 77.08 & 119.9 & 0.105 & 0.834 \\
Cancellation refund summary & 79.17 & 121.9 & 0.091 & 0.579 \\
Repeated-failure alert & 79.17 & 120.0 & 0.091 & 0.855 \\
Variant-counting skill & 72.92 & 122.3 & \textbf{0.028} & 0.509 \\
Address-history skill & 83.33 & 126.3 & 0.355 & 0.071 \\
Partial-cancellation skill & 77.08 & 125.7 & 0.052 & 0.067 \\
Bulk-request skill & 81.25 & 134.7 & 0.210 & 0.128 \\
Multi-item exchange skill & 75.00 & 126.7 & \textbf{0.029} & \textbf{0.025} \\
Numeric-option selection skill & 79.17 & 124.3 & 0.091 & 0.250 \\
Category-scope skill & 79.17 & 128.2 & 0.091 & \textbf{0.030} \\
Pending-replacement skill & 81.25 & 120.9 & 0.243 & 0.716 \\
Confirmed-action skill & 70.83 & 128.7 & \textbf{0.009} & \textbf{0.039} \\
Agent instructions & 79.17 & 119.4 & 0.091 & 0.825 \\
\end{longtable}
\endgroup

\subsection{Additional pruning results}
\label{sec:appendix-pruning-results}

We provide detailed validation results across pruning configurations, including the modules removed at each step and full-range plots of task performance and token cost. In the tables, ``protected'' denotes the configuration that retains only the protected module set.

\begingroup
\setlength{\LTcapwidth}{\linewidth}
\small
\setlength{\tabcolsep}{4pt}
\begin{longtable}{@{}p{0.13\linewidth}p{0.59\linewidth}rr@{}}
\caption{Efficiency-oriented pruning configurations for JIT agentfold on 30 DeepPlanning-Shopping validation tasks, using three trials per task. Configurations are listed in increasing order of pruning, followed by an additional all-pruned baseline. The module column lists removals relative to the preceding row. $P$ is mean match rate (\%); $T$ is mean token cost ($10^3$). Bold values mark the highest performance and lowest token cost within each table.}\label{tab:jit-pruning-b}\\
\toprule
Config. & Modules removed & $P$ & $T$ \\
\midrule
\endfirsthead
\multicolumn{4}{l}{\textit{Table \thetable{} continued}} \\
\toprule
Config. & Modules removed & $P$ & $T$ \\
\midrule
\endhead
\midrule
\multicolumn{4}{r}{\textit{Continued on next page}} \\
\endfoot
\bottomrule
\endlastfoot
$\mathcal{S}_{23}$ {\scriptsize (full)} & None & \textbf{76.10} & 329.1 \\
$\mathcal{S}_{20}$ & Shopping requirements; Shopping cost strategy; Code execution & 74.31 & 204.3 \\
$\mathcal{S}_{17}$ & Lenient JSON parsing; Tool-selection guidance; Final-answer guidance & 70.40 & 203.8 \\
$\mathcal{S}_{14}$ & Tool-list summary; Plan-execution guidance; Initial planning & 75.96 & \textbf{192.0} \\
$\mathcal{S}_{11}$ & Shopping output guidance; Compression guidance; Result-processing guidance & 73.38 & 195.0 \\
$\mathcal{S}_{8}$ & Role preamble; Planning-model guidance; General action rules & 74.10 & 215.2 \\
$\mathcal{S}_{5}$ {\scriptsize (protected)} & History folding; Action guidance; Forced final answer & 72.19 & 408.9 \\
$\mathcal{S}_{0}$ {\scriptsize (all-pruned)} & Step-output example; Workflow examples; Periodic replanning; Cart verification; Task restatement & 62.25 & 642.2 \\
\end{longtable}
\endgroup

\begingroup
\setlength{\LTcapwidth}{\linewidth}
\small
\setlength{\tabcolsep}{4pt}
\begin{longtable}{@{}p{0.13\linewidth}p{0.59\linewidth}rr@{}}
\caption{Performance-oriented pruning configurations for JIT agentfold on 30 DeepPlanning-Shopping validation tasks, using one trial per task. Configurations are listed in increasing order of pruning, followed by an additional all-pruned baseline. The module column lists removals relative to the preceding row. $P$ is mean match rate (\%); $T$ is mean token cost ($10^3$). Bold values mark the highest performance and lowest token cost within each table.}\label{tab:jit-pruning-a}\\
\toprule
Config. & Modules removed & $P$ & $T$ \\
\midrule
\endfirsthead
\multicolumn{4}{l}{\textit{Table \thetable{} continued}} \\
\toprule
Config. & Modules removed & $P$ & $T$ \\
\midrule
\endhead
\midrule
\multicolumn{4}{r}{\textit{Continued on next page}} \\
\endfoot
\bottomrule
\endlastfoot
$\mathcal{S}_{23}$ {\scriptsize (full)} & None & 75.83 & 323.7 \\
$\mathcal{S}'_{22}$ & Initial planning & 77.66 & 278.3 \\
$\mathcal{S}'_{19}$ & Plan-execution guidance; Role preamble; Result-processing guidance & \textbf{82.49} & 304.6 \\
$\mathcal{S}'_{16}$ & Compression guidance; Code execution; Shopping cost strategy & 76.88 & 219.1 \\
$\mathcal{S}'_{13}$ & Shopping output guidance; Forced final answer; Lenient JSON parsing & 72.66 & 222.8 \\
$\mathcal{S}'_{10}$ & Planning-model guidance; Shopping requirements; Final-answer guidance & 70.66 & 265.1 \\
$\mathcal{S}'_{7}$ & Tool-list summary; General action rules; Tool-selection guidance & 68.49 & 245.9 \\
$\mathcal{S}'_{4}$ {\scriptsize (protected)} & Task restatement; Periodic replanning; Cart verification & 63.28 & \textbf{145.7} \\
$\mathcal{S}_{0}$ {\scriptsize (all-pruned)} & Step-output example; Workflow examples; Action guidance; History folding & 67.99 & 664.8 \\
\end{longtable}
\endgroup

\begingroup
\setlength{\LTcapwidth}{\linewidth}
\small
\setlength{\tabcolsep}{4pt}
\begin{longtable}{@{}p{0.13\linewidth}p{0.59\linewidth}rr@{}}
\caption{Efficiency-oriented pruning configurations for PaperQA2 on LitQA2. Configurations are listed in increasing order of pruning, followed by an additional all-pruned baseline. The module column lists removals relative to the preceding row. $P$ is success rate (\%); $T$ is mean token cost ($10^3$). Results average three trials per validation task. Bold values mark the highest performance and lowest token cost within the table.}\label{tab:paperqa2-pruning-b}\\
\toprule
Config. & Modules removed & $P$ & $T$ \\
\midrule
\endfirsthead
\multicolumn{4}{l}{\textit{Table \thetable{} continued}} \\
\toprule
Config. & Modules removed & $P$ & $T$ \\
\midrule
\endhead
\midrule
\multicolumn{4}{r}{\textit{Continued on next page}} \\
\endfoot
\bottomrule
\endlastfoot
$\mathcal{S}_{18}$ {\scriptsize (full)} & None & \textbf{61.40} & 1256.6 \\
$\mathcal{S}_{15}$ & Multimodal processing; Citation metadata context; Increased evidence depth & 56.14 & 1108.4 \\
$\mathcal{S}_{12}$ & Reset tool; Structured evidence output; Answer iteration & 50.88 & 1651.2 \\
$\mathcal{S}_{9}$ & Increased concurrency; Larger text chunks; Automatic evidence gathering & 42.98 & 1728.6 \\
$\mathcal{S}_{6}$ & Evidence retrieval; Agent system prompt; Document metadata extraction & 52.63 & 2004.0 \\
$\mathcal{S}_{3}$ {\scriptsize (protected)} & Completion tool; Evidence gathering tool; Workflow guidance & 0.00 & 5915.0 \\
$\mathcal{S}_{0}$ {\scriptsize (all-pruned)} & Answer generation tool; Paper search tool; Evidence summarization & 21.93 & \textbf{0.6} \\
\end{longtable}
\endgroup

\begingroup
\setlength{\LTcapwidth}{\linewidth}
\small
\setlength{\tabcolsep}{4pt}
\begin{longtable}{@{}p{0.13\linewidth}p{0.59\linewidth}rr@{}}
\caption{Efficiency-oriented pruning configurations for OpenHands on GAIA. Configurations are listed in increasing order of pruning, followed by an additional all-pruned baseline. The module column lists removals relative to the preceding row. $P$ is success rate (\%); $T$ is mean token cost ($10^3$). Results average three trials on 60 validation tasks. Token cost uses runs with recorded usage. Bold values mark the highest performance and lowest token cost within the table.}\label{tab:gaia-pruning-b}\\
\toprule
Config. & Modules removed & $P$ & $T$ \\
\midrule
\endfirsthead
\multicolumn{4}{l}{\textit{Table \thetable{} continued}} \\
\toprule
Config. & Modules removed & $P$ & $T$ \\
\midrule
\endhead
\midrule
\multicolumn{4}{r}{\textit{Continued on next page}} \\
\endfoot
\bottomrule
\endlastfoot
$\mathcal{S}_{27}$ {\scriptsize (full)} & None & 29.44 & 799.4 \\
$\mathcal{S}_{24}$ & Version-control guidance; File editor; Skills (microagents) & 32.78 & 691.8 \\
$\mathcal{S}_{21}$ & Risk-assessment guidance; Memory guidance; Documentation guidance & 32.78 & 675.4 \\
$\mathcal{S}_{18}$ & Introductory prompt; Finish tool; External-service guidance & 33.89 & 683.3 \\
$\mathcal{S}_{15}$ & Browser tools; Pull-request guidance; Code-quality guidance & 30.56 & 654.2 \\
$\mathcal{S}_{12}$ & Troubleshooting guidance; Environment-setup guidance; Think tool & 29.44 & 663.0 \\
$\mathcal{S}_{9}$ & Efficiency guidance; Problem-solving workflow; Security guidance & 33.33 & 633.1 \\
$\mathcal{S}_{6}$ & File-system guidance; Task tracker; History condensation & 29.44 & 622.8 \\
$\mathcal{S}_{3}$ {\scriptsize (protected)} & Terminal tool; Role guidance; Process-management guidance & \textbf{35.00} & 580.3 \\
$\mathcal{S}_{0}$ {\scriptsize (all-pruned)} & Search tool; Fetch tool; Delegation & 14.44 & \textbf{5.0} \\
\end{longtable}
\endgroup

\begingroup
\setlength{\LTcapwidth}{\linewidth}
\small
\setlength{\tabcolsep}{4pt}
\begin{longtable}{@{}p{0.13\linewidth}p{0.59\linewidth}rr@{}}
\caption{Efficiency-oriented pruning configurations for LIFE-harness on $\tau^2$-bench airline. Configurations are listed from full to all-pruned. Each subsequent row removes the listed modules from the preceding configuration. $P$ is success rate (\%); $T$ is mean token cost ($10^3$). Results average three trials per validation task. Bold values mark the highest performance and lowest token cost within the table. The protected set is empty in this setting, so the configuration labeled ``protected'' is also all-pruned.}\label{tab:airline-pruning-b}\\
\toprule
Config. & Modules removed & $P$ & $T$ \\
\midrule
\endfirsthead
\multicolumn{4}{l}{\textit{Table \thetable{} continued}} \\
\toprule
Config. & Modules removed & $P$ & $T$ \\
\midrule
\endhead
\midrule
\multicolumn{4}{r}{\textit{Continued on next page}} \\
\endfoot
\bottomrule
\endlastfoot
$\mathcal{S}_{44}$ {\scriptsize (full)} & None & \textbf{83.33} & 138.7 \\
$\mathcal{S}_{42}$ & Agent instructions; Reservation flight summary & 78.33 & 124.0 \\
$\mathcal{S}_{39}$ & Updated baggage allowance; Booking hint; Booking itinerary check & 81.67 & 111.4 \\
$\mathcal{S}_{36}$ & Cancellation refund summary; Cancellation eligibility; Other-reservations skill & 80.00 & 113.2 \\
$\mathcal{S}_{33}$ & Booking payment total; Compensation amount check; Direct-flight search hint & 81.67 & 114.5 \\
$\mathcal{S}_{30}$ & Reservation payment summary; Compensation eligibility; Cancellation-check skill & 81.67 & 107.7 \\
$\mathcal{S}_{27}$ & Passenger-update hint; Delay compensation prerequisite; Route preservation & \textbf{83.33} & 116.5 \\
$\mathcal{S}_{24}$ & Connecting-flight search hint; Exact-amount skill; Basic-economy changes & 81.67 & 108.3 \\
$\mathcal{S}_{21}$ & Flight-change payment type; User reservation summary; Cabin-change eligibility & 80.00 & 115.1 \\
$\mathcal{S}_{18}$ & Baggage reduction check; Booking payment summary; Reservation-identification skill & 78.33 & 244.6 \\
$\mathcal{S}_{15}$ & Compensation hint; Booking baggage allowance; Passenger limit & 76.67 & 115.6 \\
$\mathcal{S}_{12}$ & Compensable flight status; Transfer hint; Baggage-update hint & 78.33 & 103.9 \\
$\mathcal{S}_{9}$ & Flight-change payment capacity; Cancellation hint; Passenger-count preservation & 75.00 & 90.1 \\
$\mathcal{S}_{6}$ & Flight-update payment summary; Flight-update hint; Budget-check skill & 76.67 & 99.3 \\
$\mathcal{S}_{3}$ & Certificate-use skill; Reservation policy context; Bookable flight status & 78.33 & 98.5 \\
$\mathcal{S}_{0}$ {\scriptsize (protected)} & Booking payment limits; Round-trip update skill; Repeated-failure alert & 78.33 & \textbf{87.3} \\
\end{longtable}
\endgroup

\begingroup
\setlength{\LTcapwidth}{\linewidth}
\small
\setlength{\tabcolsep}{4pt}
\begin{longtable}{@{}p{0.13\linewidth}p{0.59\linewidth}rr@{}}
\caption{Efficiency-oriented pruning configurations for LIFE-harness on $\tau^2$-bench retail. Configurations are listed in increasing order of pruning, followed by an additional all-pruned baseline. The module column lists removals relative to the preceding row. $P$ is success rate (\%); $T$ is mean token cost ($10^3$). Results average three trials per validation task. Bold values mark the highest performance and lowest token cost within the table.}\label{tab:retail-pruning-b}\\
\toprule
Config. & Modules removed & $P$ & $T$ \\
\midrule
\endfirsthead
\multicolumn{4}{l}{\textit{Table \thetable{} continued}} \\
\toprule
Config. & Modules removed & $P$ & $T$ \\
\midrule
\endhead
\midrule
\multicolumn{4}{r}{\textit{Continued on next page}} \\
\endfoot
\bottomrule
\endlastfoot
$\mathcal{S}_{48}$ {\scriptsize (full)} & None & 75.00 & 119.1 \\
$\mathcal{S}_{46}$ & Order-address hint; Return hint & 79.17 & 116.4 \\
$\mathcal{S}_{43}$ & Item availability note; Item-modification hint; Repeated-failure alert & \textbf{82.50} & 113.6 \\
$\mathcal{S}_{40}$ & Cancellation hint; Pending-replacement skill; Agent instructions & 76.67 & 106.9 \\
$\mathcal{S}_{37}$ & Return item check; Cancellation refund summary; Replacement existence check & 79.17 & 106.9 \\
$\mathcal{S}_{34}$ & Nonempty replacement list; Payment balance check; Item-modification status check & 70.00 & 110.5 \\
$\mathcal{S}_{31}$ & Exchange status check; Payment-change hint; Transfer hint & 75.83 & 106.1 \\
$\mathcal{S}_{28}$ & Numeric-option extrema; Cancellation status check; Modification product-type check & 79.17 & 103.2 \\
$\mathcal{S}_{25}$ & Numeric-option selection skill; Return refund summary; Tracking information & 77.50 & 103.6 \\
$\mathcal{S}_{22}$ & Bulk-request skill; Order eligibility note; Item-list length check & 79.17 & 105.8 \\
$\mathcal{S}_{19}$ & Address-history skill; Cancellation reason check; Nonempty item list & 75.83 & 104.7 \\
$\mathcal{S}_{16}$ & Item multiplicity check; Distinct exchange target; Partial-cancellation skill & 71.67 & 104.9 \\
$\mathcal{S}_{13}$ {\scriptsize (protected)} & Exchange item check; Order action guidance; Category-scope skill & 72.50 & 102.4 \\
$\mathcal{S}_{0}$ {\scriptsize (all-pruned)} & Multi-item exchange skill; Variant price extrema; Confirmed-action skill; Payment-change check; Exchange price difference; Exchange hint; Return eligibility check; Exchange product-type check; Profile-address hint; Available-variant count; Variant-counting skill; Distinct modification target; Modification item check & 75.83 & \textbf{100.9} \\
\end{longtable}
\endgroup

\begingroup
\setlength{\LTcapwidth}{\linewidth}
\small
\setlength{\tabcolsep}{4pt}
\begin{longtable}{@{}p{0.13\linewidth}p{0.59\linewidth}rr@{}}
\caption{Performance-oriented pruning configurations for LIFE-harness on $\tau^2$-bench retail. Configurations are listed in increasing order of pruning, followed by an additional all-pruned baseline. The module column lists removals relative to the preceding row. $P$ is success rate (\%); $T$ is mean token cost ($10^3$). Results use one trial per validation task. Bold values mark the highest performance and lowest token cost within the table.}\label{tab:retail-pruning-a}\\
\toprule
Config. & Modules removed & $P$ & $T$ \\
\midrule
\endfirsthead
\multicolumn{4}{l}{\textit{Table \thetable{} continued}} \\
\toprule
Config. & Modules removed & $P$ & $T$ \\
\midrule
\endhead
\midrule
\multicolumn{4}{r}{\textit{Continued on next page}} \\
\endfoot
\bottomrule
\endlastfoot
$\mathcal{S}_{48}$ {\scriptsize (full)} & None & 67.50 & 121.2 \\
$\mathcal{S}'_{46}$ & Replacement existence check; Payment-change hint & 77.50 & 116.3 \\
$\mathcal{S}'_{43}$ & Return hint; Nonempty replacement list; Payment balance check & 80.00 & 118.4 \\
$\mathcal{S}'_{40}$ & Item-list length check; Numeric-option extrema; Item-modification hint & 75.00 & 116.5 \\
$\mathcal{S}'_{37}$ & Return item check; Pending-replacement skill; Address-history skill & 72.50 & 117.2 \\
$\mathcal{S}'_{34}$ & Cancellation status check; Modification product-type check; Transfer hint & 70.00 & 116.1 \\
$\mathcal{S}'_{31}$ & Distinct exchange target; Bulk-request skill; Tracking information & 75.00 & 114.3 \\
$\mathcal{S}'_{28}$ & Item availability note; Cancellation reason check; Order-address hint & 75.00 & 108.4 \\
$\mathcal{S}'_{25}$ & Cancellation hint; Agent instructions; Repeated-failure alert & 65.00 & 101.4 \\
$\mathcal{S}'_{22}$ & Numeric-option selection skill; Item-modification status check; Cancellation refund summary & 70.00 & 99.9 \\
$\mathcal{S}'_{19}$ & Order eligibility note; Return refund summary; Exchange status check & 72.50 & 102.4 \\
$\mathcal{S}'_{16}$ & Available-variant count; Partial-cancellation skill; Nonempty item list & 72.50 & 106.8 \\
$\mathcal{S}'_{13}$ & Variant-counting skill; Profile-address hint; Exchange hint & 70.00 & 99.9 \\
$\mathcal{S}'_{10}$ & Return eligibility check; Exchange price difference; Payment-change check & 75.00 & 99.4 \\
$\mathcal{S}'_{7}$ {\scriptsize (protected)} & Modification item check; Exchange product-type check; Distinct modification target & 70.00 & \textbf{95.5} \\
$\mathcal{S}_{0}$ {\scriptsize (all-pruned)} & Variant price extrema; Confirmed-action skill; Multi-item exchange skill; Exchange item check; Category-scope skill; Order action guidance; Item multiplicity check & \textbf{85.00} & 100.5 \\
\end{longtable}
\endgroup

\begin{figure}[tbp]
\centering
\begin{minipage}[t]{0.495\linewidth}
\vspace{0pt}
\centering
\includegraphics[width=\linewidth]{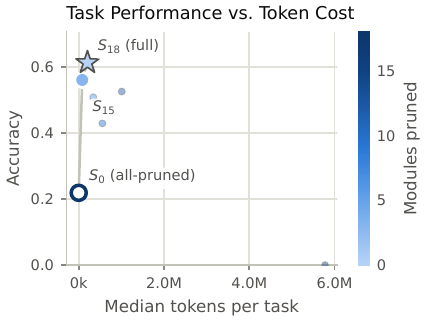}
\par {\small (a) PaperQA2 / LitQA2}
\end{minipage}\hfill
\begin{minipage}[t]{0.495\linewidth}
\vspace{0pt}
\centering
\includegraphics[width=\linewidth]{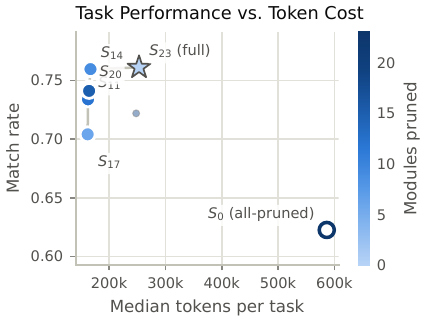}
\par {\small (b) JIT agentfold / DeepPlanning-Shopping}
\end{minipage}
\par\smallskip
\includegraphics[width=0.855\linewidth]{figures/main_results_wide/legend.pdf}
\caption{Full-range views of the PaperQA2 and JIT agentfold results in Figure~\ref{fig:main-pruning-tradeoffs}, including $\mathcal{S}_3$ and $\mathcal{S}_0$, respectively, which are omitted from the zoomed-in panels. Points average three trials; token cost is the per-trial median across tasks. Stars mark full harnesses, color indicates modules pruned, and lines trace the Pareto frontier.}
\label{fig:full-range-pruning-tradeoffs}
\end{figure}

Figure~\ref{fig:full-range-pruning-tradeoffs} provides the full-range views of the PaperQA2 and JIT agentfold plots shown in zoomed-in form in Figure~\ref{fig:main-pruning-tradeoffs}.

\subsection{Additional qualitative analysis}
\label{sec:appendix-case-studies}

We examine additional trajectories and module behavior to illustrate how pruning affects workflow progression, evidence processing, and tool use.

\paragraph{Evidence gathering and workflow guidance.} In PaperQA2, paper search adds documents to the available collection, while the evidence gathering tool converts their contents into question-relevant evidence. In an inspected trajectory without the evidence gathering tool, the agent repeatedly searches for papers without accumulating evidence contexts. After the execution budget is exhausted, fallback answer generation invokes automatic evidence gathering and produces an answer. This case shows how access to documents alone may be insufficient for the agent to progress through its workflow. Separately, removing workflow guidance, which supplies stopping rules and live progress information, is also associated with frequent budget exhaustion in the inspected runs. Both observations point to the role of workflow support in helping the agent progress and terminate.

\paragraph{Dependencies between evidence-processing modules.} PaperQA2's automatic evidence gathering runs during answer generation when no evidence contexts are available. Its workload depends on whether evidence retrieval is enabled: retrieval selects a bounded number of relevant chunks, whereas disabling it allows all available chunks to enter processing. Consequently, the cost of automatic evidence gathering depends on the functionality retained elsewhere in the harness. This code path provides a concrete example of how modules can jointly determine the work performed during execution.

\paragraph{Alternative tool use.} OpenHands exhibits shifts in tool use when particular web tools are removed. Disabling search is accompanied by more browser and fetch calls, while disabling the browser is accompanied by more search calls. When all three web tools are disabled, terminal use increases substantially. These patterns suggest that the agent attempts alternative routes when a tool becomes unavailable, redirecting activity to other tools. The availability of such alternatives can therefore influence how the agent responds to pruning.